\documentclass{article}

\usepackage[final]{corl_2021} 

\usepackage{amsmath, amssymb}
\usepackage{footmisc}
\usepackage{authblk}
\usepackage{algorithm,algorithmic}
\usepackage{mathtools}
\usepackage{subcaption}
\usepackage{xcolor}
\usepackage{tikz}
\usetikzlibrary{positioning, arrows, automata}
\usetikzlibrary{backgrounds, calc}
\usepackage{hyperref}

\usepackage{pgfplots}
\pgfplotsset{compat=1.8}
\usepackage{pgfplotstable}
\usepgfplotslibrary{groupplots}

\newcommand{\cvec}[1]{\boldsymbol{\mathrm{#1}}}

\newcommand{\KL}[2]{\textrm{KL}\left( {#1} \parallel {#2} \right)}

\newcommand{\joint}[1]{{#1}(\cvec c, \cvec \theta)}
\newcommand{\policy}[0]{\pi(\cvec \theta|\cvec c)}

\newcommand{\responsibilitypi}[0]{\pi(o|\cvec c, \cvec \theta)}
\newcommand{\tilderesppi}[0]{\Tilde{\pi}(o|\cvec c, \cvec \theta)}

\newcommand{\gatingpi}[0]{\pi(o|\cvec c)}
\newcommand{\mgating}[0]{\pi(o)}
\newcommand{\tildegating}[0]{\Tilde{\pi}(o|\cvec c)}
\newcommand{\statedistr}[0]{\pi(\cvec c)}
\newcommand{\statecomp}[0]{\pi(\cvec c|o)}

\newcommand{\comppi}[0]{\pi(\cvec \theta | \cvec c, o)}

\newcommand{\reward}[0]{\textrm{R}(\cvec c, \cvec \theta)}

\title{Specializing Versatile Skill Libraries using Local Mixture of Experts}

\author[1]{\textbf{Onur Celik}}
\author[1]{\textbf{Dongzhuoran Zhou}}
\author[1]{\textbf{Ge Li}}
\author[1]{\textbf{Philipp Becker}}
\author[1]{\textbf{Gerhard Neumann}}
\affil[1]{Autonomous Learning Robots, KIT, Germany}
\affil[ ]{\{\texttt{celik, ge.li, philipp.becker, gerhard.neumann\}@kit.edu}}
\affil[ ]{\texttt{dongzhuoran.zhou@outlook.com}}

\begin{document}
\maketitle


\begin{abstract}
A long-cherished vision in robotics is to equip robots with skills that match the versatility and precision of humans.
For example, when playing table tennis, a robot should be capable of returning the ball in various ways while precisely placing it at the desired location. 
A common approach to model such versatile behavior is to use a Mixture of Experts (MoE) model, where each expert is a contextual motion primitive.
However, learning such MoEs is challenging as most objectives force the model to cover the entire context space, which prevents specialization of the primitives resulting in rather low-quality components. 
Starting from maximum entropy reinforcement learning (RL), we decompose the objective into optimizing an individual lower bound per mixture component.
Further, we introduce a curriculum by allowing the components to focus
on a local context region, enabling the model to learn highly accurate skill representations.
To this end, we use local context distributions that are adapted jointly with the expert primitives. Our lower bound advocates an iterative addition of new components, where new components will concentrate on local context regions not covered by the current MoE.
This local and incremental learning results in a modular MoE model of high accuracy and versatility, where both properties can be scaled by adding more components on the fly. 
We demonstrate this by an extensive ablation and on two challenging simulated robot skill learning tasks. We compare our achieved performance to LaDiPS and HiREPS, a known hierarchical policy search method for learning diverse skills\footnote{Code available at \url{https://github.com/Onur4229/SVSL_LMOE}}. 
\end{abstract}



\section{Introduction}\label{sec::Introduction}
Human motor skills are precise and versatile, which allows us to perform motor tasks in different ways while achieving a consistent movement quality.
For example, when playing table tennis, we can hit the ball in various ways while still targeting a specific landing point of the ball on the opponent's side of the table. 
Another example is grasping an object which lies behind an obstacle. 
We can grasp even small objects precisely with different grasp types while avoiding the obstacle. Such versatile skills are crucial if we want to employ robots in unstructured and dynamically changing environments.
Such skills, often represented as movement primitives, were already successfully learned for challenging robot learning tasks, e.g., the ball-in-a-cup \cite{kober2014policy, klink2020self} task, by a variety of policy search algorithms \cite{deisenroth2013survey}.
Yet, most of these algorithms cannot find multiple, versatile, and precise solutions to the multi-modal solution space, as they usually assume a Gaussian policy \cite{abdolmaleki2019contextual, klink2020self, deisenroth2013survey, kupcsik2013data}. 

In this paper, we model versatile behavior with contextual skill libraries of motion primitives \cite{paraschos2013probabilistic, ijspeert2002learning}, formalized by Mixtures of Experts (MoEs).
Here the context defines task properties, e.g., reaching different goal positions or different friction parameters of an object to manipulate \cite{kupcsik2013data}. Our goal is to learn versatile skills, i.e., different movement styles to solve a given context.
Given a context, the MoE first selects a component, i.e., a motion primitive, to execute. 
Subsequently, the component adjusts the primitive's parameters and a controller executes the primitive. Such models are already commonly used \cite{daniel2012hierarchical, endLayered}.
Yet, the quality and versatility of the learned skill libraries remain limited using existing algorithms.
Most algorithms are based on expectation-maximization (EM) \cite{bishop2006pattern} techniques \cite{daniel2012hierarchical}, and suffer from local optima and mode averaging \cite{bishop2006pattern} which prevents the single components from specializing in a local context region. Moreover, existing algorithms \cite{daniel2012hierarchical, endLayered} do not explicitly optimize the versatility of the library.
Hence, they often yield degenerated libraries with only a single movement style. 
We propose a new objective for learning contextual, precise, and versatile MoE models based on a maximum entropy formulation. 
We also introduce a learnable context distribution, which provides a curriculum for each component of the MoE model. We use a variational lower bound \cite{arenz2018efficient} to decompose the objective into individual updates for the components and their related local context distributions, allowing the components to specialize in local regions of the context space and preventing mode averaging. 
Due to the curriculum, the MoE does not have to cover the whole context space during training, which prevents the averaging effects that negatively affect most other approaches.
Yet, not covering the whole context space leads to poor performance for some contexts, which is also undesirable.
Thus, we introduce a heuristic-free mechanism to add new components during training until the whole context space is covered. 
Hence our algorithm provides a modular approach that learns highly precise and versatile skills that cover the whole context space.

We evaluate our approach in a simulated beer pong and a table tennis environment. Both environments allow different motion styles to solve the tasks, which are discovered by our algorithm.
Moreover, we present ablation studies showing the importance of the single elements and hyperparameters of our algorithm.

\section{Related Work}\label{sec::RelatedWork}

{\bf Contextual Episodic Policy Search.} 
Episodic policy search \cite{deisenroth2013survey} aims at maximizing the expected return by optimizing the parameters $\cvec \theta$ of a controller, e.g., a motion primitive \cite{schaal2005dmp, paraschos2013probabilistic}. 
Most approaches use a stochastic search distribution $\pi(\cvec \theta)$ over the parameter space and aim to optimize the expected return under this search distribution \cite{deisenroth2013survey}, i.e., 
\[\max_{\pi(\cvec \theta)} \mathbb{E}_{\pi(\cvec \theta)}[R(\cvec \theta)],\]
where $R(\cvec \theta) = \sum_{t} r_t$ is the summed reward over a whole episode obtained when executing controller parameter $\cvec \theta$. As it is common in the literature, we will denote $\pi(\cvec \theta)$ as our policy even though it only indirectly chooses the control actions of the agent by selecting the controller parameters $\cvec \theta$.
Different optimization schemes such as policy gradients \citet{SehnkeParamExploring}, natural gradients  \cite{wierstra2014natural}, stochastic search strategies  \cite{hansen2001completely, mannor2003cross} or trust-region optimization techniques \cite{daniel2012hierarchical,abdolmaleki2016model} have been used. 
Researchers extended these approaches to the contextual setting \cite{tangkaratt2017policy, abdolmaleki2019contextual}, where the search distribution  $\pi(\cvec \theta|\cvec c)$ now depends on a context vector $\cvec c$ which describes the task, e.g., a goal location to reach. The contextual objective is typically formalized as \[\max_{\pi(\cvec \theta|\cvec c)} \mathbb{E}_{p(\cvec c)}\left[\mathbb{E}_{\pi(\cvec \theta|\cvec c)}[R(\cvec \theta, \cvec c)] \right],\] where $p(\cvec c)$ is the given distribution over context vector and the rewards now also depends on the context. 
\citet{klink2020self} introduce a curriculum into contextual policy search.
By having an adaptable distribution over the contexts, they allow the agent to concentrate on easy-to-solve contexts first and then generalize to the whole context space.

{\bf Versatile Skill Learning.} The Hierarchical Relative Entropy Policy Search (HiREPS) algorithm \cite{daniel2012hierarchical} extends the classical Relative Entropy Policy Search (REPS) approach \cite{peters2010relative} to MoEs, which allows learning versatile skills in a contextual episodic policy search setting.
In a similar approach, Layered Direct Policy Search (LaDiPS) \cite{endLayered} also uses MoE policies, but builds on Model-Based Relative Entropy Policy Search (MORE) \cite{abdolmaleki2016model} instead of REPS.
Both HiREPS and LaDiPS address the same problems as our approach, yet there are also considerable differences.
First, HiREPS jointly optimizes the whole mixture model and introduces an additional constraint, which bounds the entropy loss of the responsibilities in each iteration. This constraint is crucial for obtaining versatile and well-performing solutions.
LaDiPS uses separate updates for the different parts of the mixture but also relies on additional constraints, where the entropy of the gating is lower bounded with a constant value.
In contrast, for our approach, the objective and separate updates of the individual mixture parts follow naturally from the maximum entropy formulation.
Second, neither HiREPS nor LaDiPS uses a curriculum for training.
Thus, in both approaches, the MoE always has to cover the whole context space and, hence, the components cannot specialize. 

{\bf Variational Inference.} 
Our work is also related to several recent advances in variational inference  \cite{arenz2018efficient, arenz2020trust, Becker2020Expected}.
It is well known that maximum entropy RL is equivalent to inference in an appropriate probabilistic model \cite{levine2018reinforcement}. 
Similar to previous works~\cite{neumann2011variational, haarnoja2018sac}, we exploit this relation and draw inspiration from recent research into variational inference and density estimation for Gaussian mixture models and MoEs \cite{arenz2018efficient, arenz2020trust, Becker2020Expected}. 
We reformulate the lower bound objectives introduced in those approaches for our maximum entropy RL setting and extend them with a curriculum for the mixture components.  

{\bf Related Step-Based Approaches.}
In the step-based setting, the policy does not learn a function from contexts to parameters of an episodic controller but directly maps from system states to actions and the policy updates are performed with the information from each time-step.
Practitioners often use deep neural networks to parameterize step-based policies, giving rise to the field of deep RL.
In this context, versatile policy learning is also a very active research area \cite{eysenDIAYN, kumar2020one, osa2021discovering, campos2020explore}. 
These approaches use a similar MoE model where the mixture component is only chosen at the beginning of an episode. Yet, the component is chosen randomly without conditioning on a context or state variable. Moreover, these approaches reformulate a mutual information based objective into a maximum entropy objective while we develop a more direct maximum entropy maximization.

{\bf Curriculum Learning.}
Researchers also worked on introducing curricula into deep RL. 
In a first approach \citet{ghosh2018divideandconquer} proposed partitioning the initial state distribution using clustering. 
They then learn individual policies for each partition while keeping the partitioning fixed.
Strictly, this is not a curriculum as the partitioning is not adjusted, yet it still allows specialization of the individual policies in different regions of the state space. 
To automatically generate and adapt a curriculum for deep RL approaches, \citet{Klink2020crl} extended their approach from the episodic setting \cite{klink2020self} to the step-based setting. 
Yet, neither of these approaches addresses versatility. While we follow an episode-based approach, both methodologies have their benefits and limitations \cite{deisenroth2013survey} which are, however, not the focus of this paper. We offer further discussion and quantitative comparison to a step-based approach for a common benchmark task in the appendix, therefore.

{\bf Options.}
A related hierarchical approach is the options framework \cite{sutton1999between, bacon2017option, riemer2018learning}. The options framework extends the standard MDP to a semi-MDP to include a temporal abstraction of low-level control policies. Given a termination condition, the executed low-level policy can be terminated and another can be turned on. Our policy structure can be seen as a simplification of the options framework where the option is only selected at the beginning of each episode. Yet, the options framework does not explicitly address learning versatile skills.
\section{Specializing Versatile Mixture of Expert Models}\label{sec::SpecializingVersatileMixtureOfExpertModels}
To allow versatile solutions, we employ a Mixture of Experts (MoE) model as policy representation which is given as $\policy=\sum_o \gatingpi \comppi$, where $\gatingpi$ is the gating distribution, assigning a probability to component $o$ given the context $\cvec c$ and $\comppi$ is the component distribution for component $o$, which adapts the motion primitive's parameters $\cvec \theta$ to the given context $\cvec c$. 
In this section, we derive a lower bound to optimize each component and its corresponding context distribution independently.
In order to implement a curriculum over the context $\cvec c$, we also introduce a learned context distribution $\pi(\cvec c) = \sum_o \statecomp \mgating$, which is also a mixture model specified by the component-wise context distribution $\statecomp$ and the component weights $\mgating$.   By applying Bayes' Rule to replace the gating distribution  $\gatingpi$, we can now rewrite the general mixture of experts model as
\vspace{-0.15cm}
\begin{align}\label{eq::MixtureModel}
    \policy = \sum_o \frac{\statecomp \mgating}{\statedistr}\comppi.
\vspace{-0.15cm}
\end{align}
This policy definition allows each component $o$ to adjust its curriculum by explicitly optimizing for $\statecomp$ (Section \ref{subsec::LBCtxtDistrs}) and thus concentrating on a local region in the context space. We model the components $\comppi$ as a linear conditional Gaussian distribution and the component-wise state distribution $\statecomp$ as a Gaussian. The prior weights $\mgating$ define a categorical distribution. Throughout the next sections, we denote every probability distribution which is adaptable through the optimization process with $\pi$ as it is part of our policy $ \policy$. Furthermore we show the full derivations for the next sections in the Appendix \ref{sec::app::derivations}.

\subsection{Maximum Entropy Skill Learning with Curriculum}\label{subsec::MaxEntrRlWithCurr}

We consider a maximum entropy objective \cite{levine2018reinforcement} for episodic policy search, i.e., 
\begin{align}\label{eq::StandardMaxEntrObj}
    \max_{\policy} \mathbb{E}_{p(\cvec c)}\left[\mathbb{E}_{\policy} \left[\reward\right] + \alpha \textrm{H}\left[\policy\right]\right],
\end{align}
where $p(\cvec c)$ is the task specific context distribution, $\reward$ is the reward function,  $\textrm{H}(\policy) = -\int_{\cvec \theta}\policy \log \policy d\cvec \theta$ the entropy and $\policy$ is our MoE model. The reward maximization enforces preciseness while the entropy bonus enforces versatility. However, the standard maximum entropy objective does not allow each mixture component to create its own curriculum, since an optimization over the component-wise context distributions $\statecomp$ is not given.
Inspired by the work from \citet{klink2020self}, we extend and modify the objective to
\begin{align}
\label{eq::maxEntrObj}
    \max_{\policy,\pi(\cvec c)} \mathbb{E}_{\pi(\cvec c)}\left[\mathbb{E}_{\policy} \left[\reward\right] + \alpha \textrm{H}\left[\policy\right]\right] - \beta\KL{\statedistr}{p(\cvec c)},
\end{align}
where $\alpha $ and $\beta $ are scaling parameters, $\textrm{H}(\policy) = -\int_{\cvec \theta}\policy \log \policy d\cvec \theta$ is the entropy and $\KL{\statedistr}{p(\cvec c)}=\int_{\cvec c} \statedistr \log \frac{\statedistr}{p(\cvec c)} d\cvec c$ denotes the KL-divergence. Note the difference in the optimization variables compared to Eq. (\ref{eq::StandardMaxEntrObj}). The Kullback-Leibler (KL) term ensures that the context distribution $\statedistr$ is close to the task specific context distribution $p(\cvec c)$ while $\statedistr$ can choose to have low probability in regions of the context space where the MoE model is performing poorly.  
Note that this objective is similar to the negative  I-projection of the joint distribution $\pi(\cvec c, \cvec \theta)$ used in variational inference. Here, we also exploit the properties of the I-projection for learning the context distribution -- the I-projection is mode seeking instead of mode-averaging and therefore allows for specialization on a local context area. 
Yet, the given objective is difficult to optimize for mixture models as the sum over the mixture components is appearing inside the log terms of the entropy and the KL. However, similar to \cite{arenz2018efficient}, we can replace $\policy$ in Obj. (\ref{eq::maxEntrObj}) with our mixture model and apply Bayes theorem to arrive at
\begin{align}\label{eq::ModelOBJ}
    \max_{\joint{\pi}} & \mathbb{E}_{\pi(o),\statecomp}\Big[\mathbb{E}_{\comppi} \overbrace{\big[\reward   
     + \alpha \log \responsibilitypi \big]}^{\textrm{augmented reward for component $o$}} + \overbrace{\beta\log p(\cvec c) + (\beta-\alpha) \log \gatingpi }^{\textrm{augmented reward for context distributions}}\Big] \\& + \alpha \mathbb{E}_{\mgating,\statecomp}\big[\textrm{H}\left[\comppi\right]\big]
   + \quad \beta \mathbb{E}_{\mgating}\big[\textrm{H}\left[\statecomp\right]\big] + \quad \beta\textrm{H}\left[\mgating\right]. \nonumber
\end{align}
 The exact derivations are given in the Appendix \ref{sec::app::derivations}. Note that Eq. (\ref{eq::ModelOBJ}) is equivalent to Eq. (\ref{eq::maxEntrObj}), yet, instead of the entropy for the whole mixture model, it now contains entropy terms for each hierarchy layer of the MoE model, i.e., $\textrm{H}\left[\comppi\right]$, $\textrm{H}\left[\statecomp\right]$ and $\textrm{H}\left[\mgating\right]$, which are much simpler to compute. 
We also introduced log responsibilities $ \responsibilitypi =  \comppi \gatingpi/\policy$ and $ \gatingpi= \statecomp\mgating/\statedistr$, occurring in the augmented rewards. They return a high negative reward for component $o$, if the context-parameter pair $(\cvec c, \cvec \theta)$ or the context sample $\cvec c$ is already covered by another component, pushing the component to uncovered regions of the parameter space or context space respectively. Yet, the regions for the components will still overlap due the entropy bonuses for $\comppi$ and $\statecomp$. Without this reward augmentation, each component could be optimized completely independently in Eq. (\ref{eq::ModelOBJ}) using a max-entropy objective. However, in this case, all components would concentrate on learning the best solution irrespective of whether this solution has already been covered by another component.  
Yet, the log responsibilities still hinder us from optimizing each component $\comppi$ and its corresponding context distribution $\statecomp$ independently, since the sum over o from the mixture models $\statedistr$, $\policy$ respectively appears in the log term. In the following sections we show, how we can overcome this limitation by introducing a lower bound inspired by variational inference \cite{arenz2018efficient}. As we consider each possible context as equally important, $p(\cvec c)$ is assumed uniformly distributed in a given interval in the following and thus, can be neglected in the objective. 
\subsection{Lower-Bound Decomposition for Component Distributions}\label{subsec::LBCompDists}
In order to maximize the objective in Eq. (\ref{eq::ModelOBJ}) for each component $\comppi$ individually, we can first extract the terms which only depend on $\comppi$ for a specific $o$ as
\begin{align}\label{eq::LBComps_without_aux}
    \max_{\comppi} & \mathbb{E}_{\statecomp,  \comppi} \left[\reward + \alpha \log \responsibilitypi \right] +\alpha\mathbb{E}_{\statecomp}\left[\textrm{H}\left[\comppi\right]\right].
\end{align}
The responsibilities are still hindering us to optimize each component $\comppi$ independently. However, similar to \cite{arenz2018efficient}, we can obtain a tight lower-bound by introducing a variational distribution $\tilderesppi$ and replacing the responsibilities in Eq. (\ref{eq::LBComps_without_aux}) with $\tilderesppi$.
This variational distribution is fixed during the optimization and can be computed according to the last policy model allowing us to update each component independently. It is easy to show that after the update of the variational distribution $\tilderesppi=\pi_{\textrm{old}}(o|\cvec c, \cvec \theta)$, the introduced lower bound is tight. Please refer to the appendix.  
The resulting lower bound is a standard maximum entropy RL objective with an additional reward augmentation of $\alpha \log \tilderesppi$  and thus can be optimized with any suitable Policy Search algorithm. Here we use a maximum-entropy-adjusted version of contextual MORE \cite{tangkaratt2017policy}.

\subsection{Lower-Bound Decomposition for Context Distributions and Prior Weights}\label{subsec::LBCtxtDistrs}
To update $\statecomp$ for a specific $o$, we extract the relevant terms from Obj. (\ref{eq::ModelOBJ}) 
\begin{align}\label{eq::LBCtxtDistr_with_aux}
    \max_{\statecomp}&  \mathbb{E}_{\statecomp} \left[L_c(o,\cvec c) + (\beta - \alpha) \log \tildegating\right] + \beta \textrm{H}\left(\statecomp\right),
\end{align}
where  $ L_c(o, \cvec c) = \mathbb{E}_{\comppi} \big[\reward   + \alpha \log \tilderesppi\big] + \alpha \textrm{H}\left[\comppi\right] $ corresponds to the expected augmented maximum entropy objective of component $\comppi$ in context $\cvec c$  and $\tildegating=\pi_{\textrm{old}}(o|\cvec c)$ is a second variational distribution, which we introduced to be able to optimize for each $\statecomp$ individually. Similarly to the previous section, the objective given in Eq. (\ref{eq::LBCtxtDistr_with_aux}) is a tight lower bound to the original objective where the responsibilities $\gatingpi$ are used instead of $\tildegating$.
We approximate the integral over $\cvec \theta$ with a single sample, as we typically only have a single parameter sample per context available. Yet, as we still have the outer expectation $\mathbb{E}_{\statecomp}$ in Eq. (\ref{eq::LBCtxtDistr_with_aux}) which we approximate by multiple context samples, the whole Monte-Carlo estimation of the expectations is still unbiased and with low variance.
After the optimization step (Eq. \ref{eq::LBCtxtDistr_with_aux}), we obtain the optimal solution $\pi^*(\cvec c|o)$ and tighten the bound by setting $\tildegating=\pi^*(o|\cvec c)$ and $\tilderesppi = \pi^*(o|\cvec c, \cvec \theta)$.
Again, this lower bound corresponds to an augmented maximum entropy RL objective and thus we can be updated it with any suitable policy search method. Like \cite{arenz2018efficient}, we use an adjusted version of MORE \cite{abdolmaleki2016model}.

Finally, we can formulate the objective for updating the component weights $\mgating$, which resembles a lower bound of the original Objective (\ref{eq::ModelOBJ}) and corresponds to the highest hierarchy in our update scheme. The objective is given as
\begin{align}\label{eq::OBJWeights}
    \max_{\mgating} &\sum_o \pi(o) \left[\mathbb{E}_{\statecomp}\left[L_c(o,\cvec c) + (\beta - \alpha) \log \tildegating\right] + \beta \textrm{H}\left(\statecomp\right)\right] + \beta \textrm{H}\left(\mgating\right),
\end{align}
which is a maximum entropy RL objective for categorical distributions. Here, we use REPS \cite{peters2010relative}. 

To summarize, we split the initial objective in Eq. (\ref{eq::ModelOBJ}) into different hierarchies, allowing us to optimize the different terms in our mixture model individually. Starting by first updating the components $\comppi$ using the maximization problem in Eq. (\ref{eq::LBComps_without_aux}), we can optimize for the component-wise context distributions $\statecomp$ with Obj. (\ref{eq::LBCtxtDistr_with_aux}) after tightening the bound. The components $\comppi$ adjust the movement primitive parameters $\cvec \theta$ given a context $\cvec c$, while the component-wise context distributions $\statecomp$ ensure that the components only see context samples from a local context region. Finally, we update the weight distribution $\mgating$ using Eq. (\ref{eq::OBJWeights}). 
\paragraph{Algorithmic Details and Addition of Components.}
We initialize our algorithm with only one component and incrementally add components, and their corresponding context distributions, randomly. We fix all components except for the newly added one and optimize it for $K$ iterations to let it discover new solutions in yet undiscovered context regions. For updating $\comppi$ (Obj. (\ref{eq::LBComps_without_aux})) we sample from the local context distribution $\statecomp$. By also updating $\statecomp$ according to Obj. (\ref{eq::LBCtxtDistr_with_aux}), the components can adjust their curriculum and search for their favored context regions.
After $K$ iterations we add a new component and repeat the procedure. Due to the augmented rewards, the new component will focus on undiscovered solutions and the local context distribution will cover uncovered areas of the context space. Note that such a simple adding procedure is only possible due to the mode-seeking properties of the I-projection, as the components do not need to average over multiple modes but can specialize on a local context region.  
We also fix the weights $\mgating$ to be uniformly distributed among all components during learning, since otherwise components which are already fully trained might dominate the optimization. By allowing to fine-tune all components every $H$ iterations, the previously added components can adjust to the newly added ones. After finishing adding components, we update the weights $\mgating$ at the end of our optimization procedure. The variables $K$ and the number of components added in total are task-dependent. We describe the algorithm in more detail in the Appendix \ref{sec::app:alg_details}.
\begin{figure}[t!]
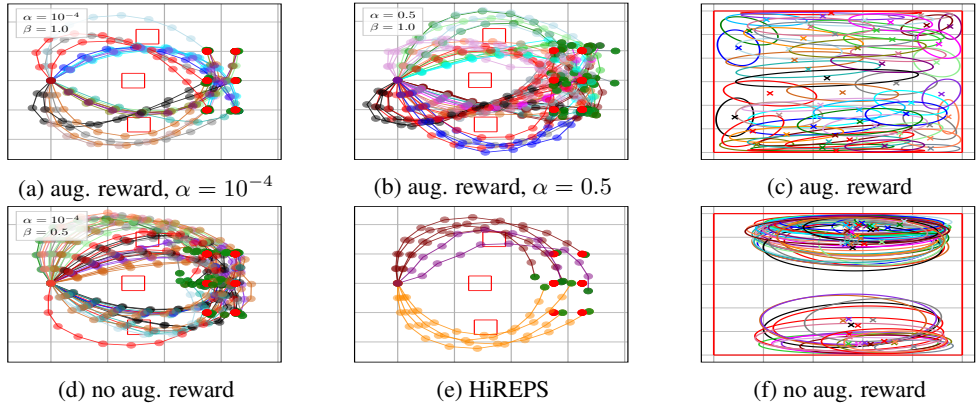

\centering
    \begin{minipage}[t]{0.33\textwidth}
        \centering
        \resizebox{0.8\textwidth}{0.45\textwidth}{\input{plots/ablation2d/with_resp/w_resp_a_0.0001_b_1}}
       \subcaption[]{\centering  aug. reward, $\alpha = 10^{-4}$} 
        \label{fig::abl::wra0.0001b1}
    \end{minipage}%
    \begin{minipage}[t]{0.33\textwidth}
        \centering
       \resizebox{0.8\textwidth}{0.45237\textwidth}{\input{plots/ablation2d/with_resp/w_resp_a_0.5_b_1}}
       \subcaption[]{\centering  aug. reward, $\alpha = 0.5$}
       \label{fig::abl::wra0.5b1}
    \end{minipage}%
    \begin{minipage}[t]{0.33\textwidth}
        \centering
       \resizebox{0.8\textwidth}{0.45\textwidth}{\input{plots/ablation2d/with_resp/ctxt_distr_w_resp_a_0.0001_b_1}}
       \subcaption[]{\centering  aug. reward} 
       \label{fig::abl::ctxt_distr_wra0.0001b1}
    \end{minipage}
    \begin{minipage}[t]{0.33\textwidth}
        \centering
        \resizebox{0.8\textwidth}{0.45\textwidth}{\input{plots/ablation2d/without_resp/wo_resp_a_0.0001_b_0.5}}
        \subcaption[]{\centering no aug. reward} 
        \label{fig::abl::wora0.0001b0.5}
    \end{minipage}%
    \begin{minipage}[t]{0.33\textwidth}
        \centering
        \resizebox{0.8\textwidth}{0.45\textwidth}{\input{plots/ablation2d/hireps_best_model_e0.5_k0.99}}
        \subcaption[]{HiREPS} 
        \label{fig::abl::hireps}
    \end{minipage}%
    \begin{minipage}[t]{0.33\textwidth}
        \centering
         \resizebox{0.8\textwidth}{0.45\textwidth}{\input{plots/ablation2d/without_resp/ctxt_distr_wo_resp_a_0.0001_b_0.5}}
        \subcaption[]{\centering no aug. reward}
       \label{fig::abl::ctxt_distr_wora0.0001b0.5}
    \end{minipage}%
    \vspace{-2mm}
    \caption{\textbf{Importance of the responsibilities in the augmented rewards and comparison to HiREPS.} A snapshot of the learned policies of the planar reaching task for a 2-dim context space (illustrated in (c) and (f)) by considering $\log \tilderesppi$, $\log \tildegating $ ((a)+(b)+(c)) and neglecting the auxiliary distributions ((d)+(f)). In ((a)+(b)+(c)) we can learn more diverse solutions (a) + (b)  and cover the whole 2d-context space (c) with different entropy bonuses, whereas the solutions in ((d)+(f)) show nearly the same, partially invalid solutions by going through the red rectangles (obstacles) (d) and are not able to cover the whole context space (f) leading to poor generalization performance. The solutions by HiREPS (e) indicate less versatility compared to the solutions (a) + (b). Note that each color indicates a different component and the red dots indicate the 6 chosen context vectors used for sampling.}
	\label{fig::Ablation_PlanarRobot}
	\vspace{-2.0mm}
\end{figure}

\section{Experiments}\label{sec::Experiments}

We start by investigating the importance of the different terms of the objectives derived in Section \ref{sec::SpecializingVersatileMixtureOfExpertModels} and subsequently evaluate the versatility and precision of the learned skills on simulated robot beer pong and table tennis experiments. In Appendix \ref{sec::app:exp_details} we report all hyperparameters and provide a comparison and analysis to a step-based policy search strategy (PPO \cite{schulman2017proximal}) in different scenarios.

\subsection{Ablation Studies}\label{sec::Ablation}
We investigate the importance of the augmented rewards on a 10-link planar reaching task with two-dimensional context space. For this purpose, we update the component distributions $\comppi$ and the corresponding context distributions $\statecomp$ by optimizing the objectives in (\ref{eq::LBComps_without_aux}, \ref{eq::LBCtxtDistr_with_aux}) with i) considering the responsibilities (as given by our algorithm) and ii) by setting $\log \tilderesppi$ and $ \log \tildegating$ to zero. 
Furthermore, we compare the solutions found by our method to the solutions found by the SOTA method HiREPS. 
\begin{figure}[t!]
    \begin{minipage}[b]{0.25\textwidth}
        \centering
       \resizebox{0.97\textwidth}{!}{\input{plots/experiments/episodic_bp/reward_functions_with_ladips_2times_standard_error}}
       \subcaption[]{Rewards-{\bf BP}}
       \label{fig::exps_bp_episodic::rewards}
   \end{minipage}\hfill
    \begin{minipage}[b]{0.25\textwidth}
        \centering
       \resizebox{0.925\textwidth}{!}{\input{plots/experiments/episodic_tt_small_range/rewards_with_ladips}}
       \subcaption[]{Rewards-{\bf TT}}
       \label{fig::exps_tt_episodic::rewards}
   \end{minipage}\hfill%
           \begin{minipage}[b]{0.25\textwidth}
       \centering
       \resizebox{\textwidth}{!}{\input{plots/experiments/episodic_bp/cmp_ctxt_bp}}
       \subcaption[]{Context-Space {\bf BP}}
       \label{fig::exps:bp_episodic::ctxt_distr}
   \end{minipage}\hfill%
   \begin{minipage}[b]{0.25\textwidth}
       \centering
       \resizebox{\textwidth}{!}{\input{plots/experiments/episodic_tt_small_range/ctxt_distr_tt_small_range}}
       \subcaption[]{Context-Space {\bf TT}}
       \label{fig::exps:tt_episodic::ctxt_distr}
   \end{minipage}

   \begin{minipage}[b]{0.25\textwidth}
        \centering
        \resizebox{\textwidth}{!}{\input{plots/experiments/episodic_bp/svsl_heat_map}}
        \subcaption[]{Successes {\bf BP} Ours}
        \label{fig::exps_bp_episodic::svsl_heat_map}
    \end{minipage}\hfill%
    \begin{minipage}[b]{0.25\textwidth}
        \centering
       \resizebox{\textwidth}{!}{\input{plots/experiments/episodic_bp/hireps_heat_map}}
       \subcaption[]{Successes {\bf BP} HiREPS}
       \label{fig::exps_bp_episodic::hireps_heat_map}
   \end{minipage}\hfill%
   \begin{minipage}[b]{0.25\textwidth}
       \centering
       \resizebox{\textwidth}{!}{\input{plots/experiments/episodic_bp/svsl_z_axis_ball_trajs}}
       \subcaption[]{Ball Trajectories {\bf BP}}
       \label{fig::exps_bp_episodic::ball_trajs}
   \end{minipage}\hfill%
   \hspace{2mm}
   \begin{minipage}[b]{0.2\textwidth}
       \centering
       \includegraphics[height=2.32cm, width=2.7cm]{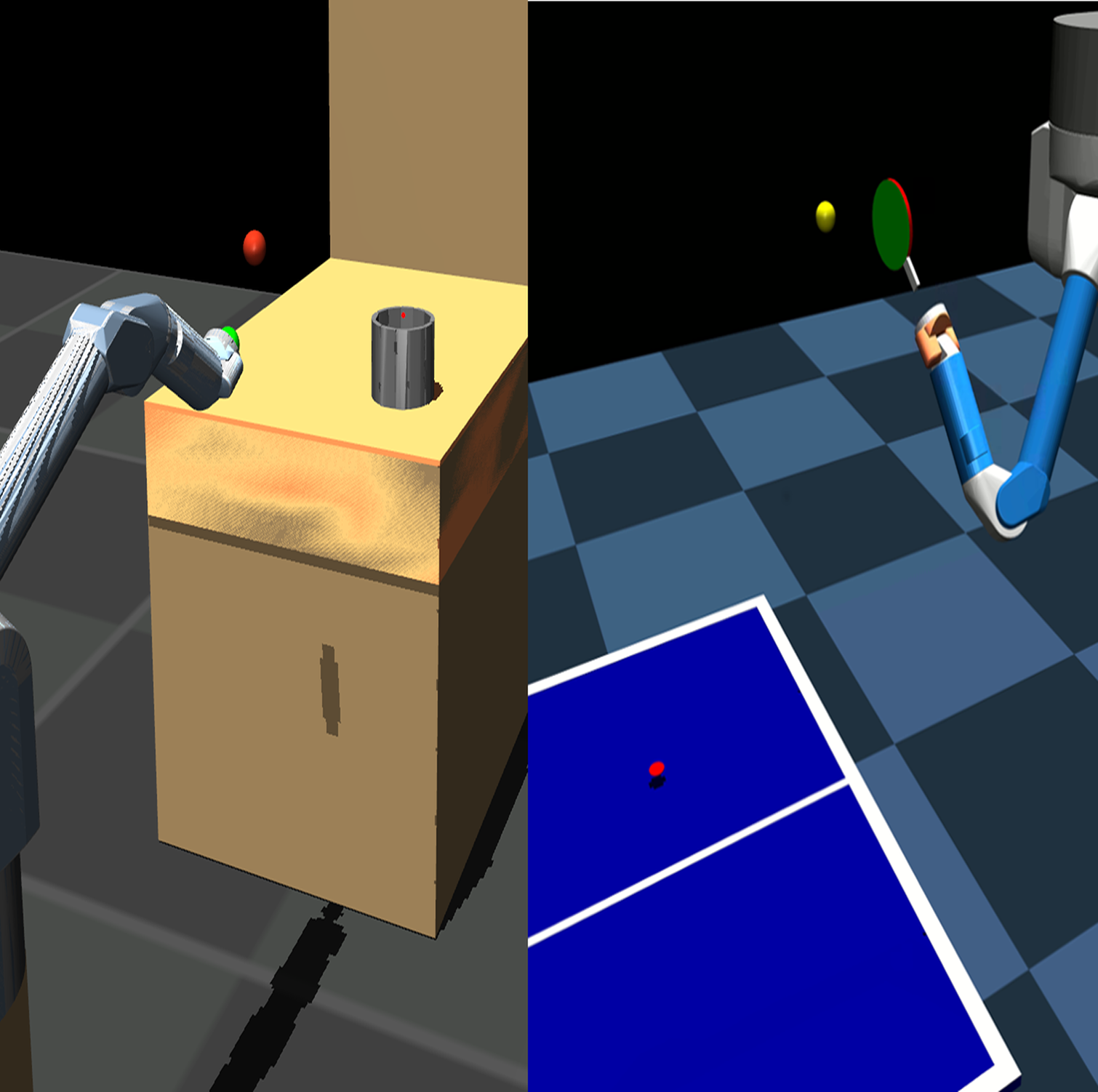}
       \vspace{4.7mm}
       \subcaption[]{Environments}
       \label{fig::exps:bp_tt_setting}
   \end{minipage}\hfill
   \caption{\textbf{Beer Pong (BP) and Table Tennis (TT) Experiments.}  In the {\bf BP} experiment --left of (h)-- the robot has to throw the red ball into a cup on the table. In the {\bf TT} experiment --right of (h) -- the robot hast to return the ball to the opponent's table side. The learned mixture of expert (MoE) models show high quality of the skills, reflected in the high reward values (a + b) and dense coverage of the context spaces, i.e., the 2-dim. position of the cup (c) for {\bf BP} and the 4-dim desired outgoing landing and incoming landing ball positions (d) for {\bf TT}. For {\bf BP} we also compare the success rates (92\% in average) of our approach (e) and HiREPS (75\% in average) (f) throughout the context space, where we consider a trial successful if the ball goes into the cup. 
   For {\bf BP} versatile skills induce versatile ball trajectories for a given context (g). Versatile strikes for the {\bf TT} experiment are shown in Fig. \ref{fig::exps_tt_episodic::versatile_strikes})}
	\vspace{-1.7mm}
\end{figure}

{\bf Versatility is Induced by the Augmented Rewards.} In the 10-link planar reaching task (see Figure \ref{fig::abl::wra0.0001b1}) the robot has to reach the red dots with its end-effector in a 2-dim context-space, while avoiding the rectangle-shaped obstacles. By adding 60 components, we have trained both versions (i and ii) over 15 seeds/trials and have chosen the best performing parameter constellations (Fig. \ref{fig::abl::wra0.0001b1}) for i), Fig. \ref{fig::abl::wora0.0001b0.5}) for ii)).  We then picked the first model and sampled for each of the shown six contexts vectors (red dots) 100 samples and plotted the corresponding mean of each sampled component. The plots for i) (Fig. \ref{fig::abl::wra0.0001b1}) + Fig. \ref{fig::abl::wra0.5b1})) show versatile -- several modes to the same context -- and precise -- small distance of end-effector (green dot) to goal (red dot), while avoiding obstacles-- solutions while fully covering the context space (Fig. \ref{fig::abl::ctxt_distr_wra0.0001b1}). A trend of covering more modes with higher $\alpha$ can also be seen in (Fig. \ref{fig::abl::wra0.5b1})). For the case without the responsibilities ii) (Fig. \ref{fig::abl::wora0.0001b0.5})), the solutions are not precise and invalid --reaching through the obstacles--. As Fig. \ref{fig::abl::ctxt_distr_wora0.0001b0.5})) shows, the context distributions focus on easy-to-solve context regions (top and bottom part) and do not cover the full context space, leading to extrapolated solutions from components that are not trained for these contexts (red dots).

{\bf Comparison to HiREPS.} We pick the best performing model among 15 seeds/trials after hyperparameter optimization. In Fig. \ref{fig::abl::hireps}) the mean solutions of the sampled components --sampled in the same way as before -- can be seen. HiREPS shows less diverse and qualitative solutions, where only 3 of the 60 components were chosen in total.

\subsection{Simulated Robotic Experiments}
We test and compare our algorithm on simulated Beer Pong and Table Tennis environments in MuJoCo \cite{emo}, where we have chosen HiREPS \cite{daniel2012hierarchical} and LaDiPS \cite{endLayered} as baselines. Both methods are suitable baselines since they are contextual policy search algorithms and consider optimizing a mixture model. Throughout our experiments we choose probabilistic movement primitives (ProMP) \cite{paraschos2013probabilistic} as policy representation, where depending on the weights $\cvec \theta $ desired trajectories are generated and subsequently tracked with a PD-controller. In our experiments, given a context $\cvec c$, the components $\comppi$ adjust the weight vector and the length of the trajectory, which are summarized in the vector $\boldsymbol{\theta}$. 
We consider non-markovian rewards, in which the reward depends on the history of state and actions. This type of reward function is not applicable to common step-based RL methods which build on markovian properties. 
As for analyzing the results of the experiments, we focus on the questions i) how does our algorithm perform compared to SOTA baselines, ii) are we able to cover the whole context space, and iii) are we able to learn versatile skills?

{\bf Beer Pong.} The goal of the Barret WAM robot is to throw the red ball into the cup on the table. The 2-dim contexts resemble the position of the cup on the table. 
We incrementally add 70 components in our method, while HiREPS and LaDiPS directly start with 70 components. We have run all algorithms over 20 seeds/trials and report the performance in Fig. \ref{fig::exps_bp_episodic::rewards})\footnote{\label{exclude_last_comp}To reflect the model's true performance, the lastly added component was excluded from testing, since it is not fully trained yet and would not be chosen by the model if $\mgating $ would not be a uniform-distribution.}, where we plot the mean reward with two times the standard error. While HiREPS already converges after around half a million samples, our approach can quickly outperform it. Since LaDiPS uses intra-option learning, it outperforms our method in terms of sample efficiency. However, with the increasing number of components, we achieve a higher end-reward indicating that we can learn more qualitative solutions. 
Our algorithm allows to cover the whole context space with the learned component-wise distributions (see Fig. \ref{fig::exps:bp_episodic::ctxt_distr})) resulting in a high end-reward. This performance is reflected in Fig. \ref{fig::exps_bp_episodic::svsl_heat_map}), where we have divided the context space into a fine grid and sampled 100 times for each context a component from which we have executed the mean. We repeated that for all 20 different models and plot the mean success rate of throwing the ball into the cup. We did the same procedure for HiREPS in Fig. \ref{fig::exps_bp_episodic::hireps_heat_map}). We can observe that HiREPS has much darker rectangles, showing that the success rate in these context regions is low. Although versatility is encouraged in the joint space of the robot, different joint trajectories often yield different ball trajectories. Given one context, in Fig. \ref{fig::exps_bp_episodic::ball_trajs}) we show the z-coordinates of the ball trajectories over time resulting from sampling 20 times from the MoE model. Each component leads to a different number of "ball-jumps". In Appendix \ref{sec::apdx::VersatilityBP} we qualitatively show that we can learn more versatile solutions than LaDiPS and report that we can achieve a much higher expected mixture entropy (Fig. \ref{fig::exps_bp_episodic::apdx:mixt_entr}), which indicates that our solutions are versatile.

\begin{figure}[t!]	
  \begin{minipage}[t!]{\textwidth}
    \begin{minipage}[t!]{\textwidth}
	\begin{minipage}[t!]{0.115\textwidth}
		\includegraphics[width=\textwidth]{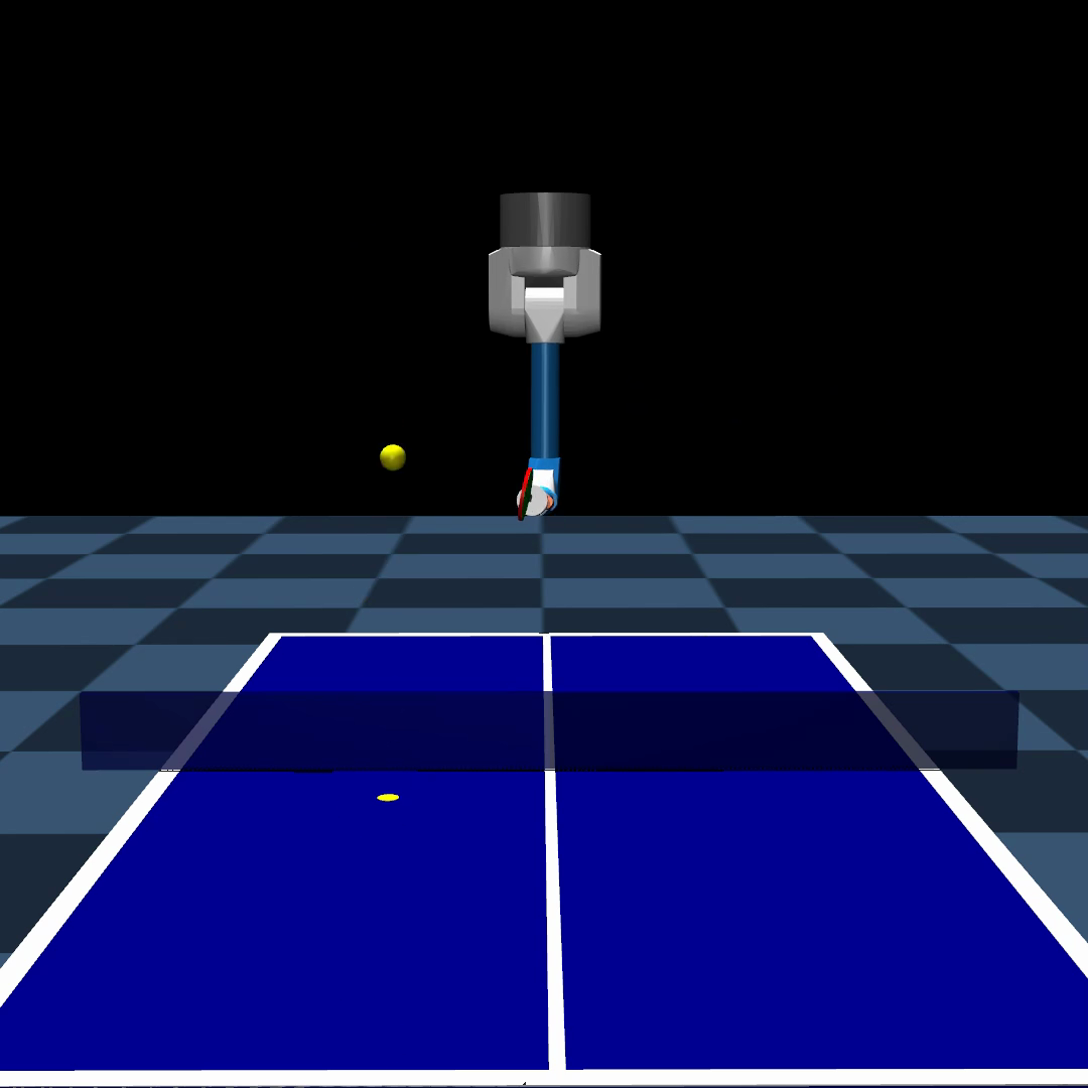}
	\end{minipage}\hfill
	\begin{minipage}[t!]{0.115\textwidth}
		\includegraphics[width=\textwidth]{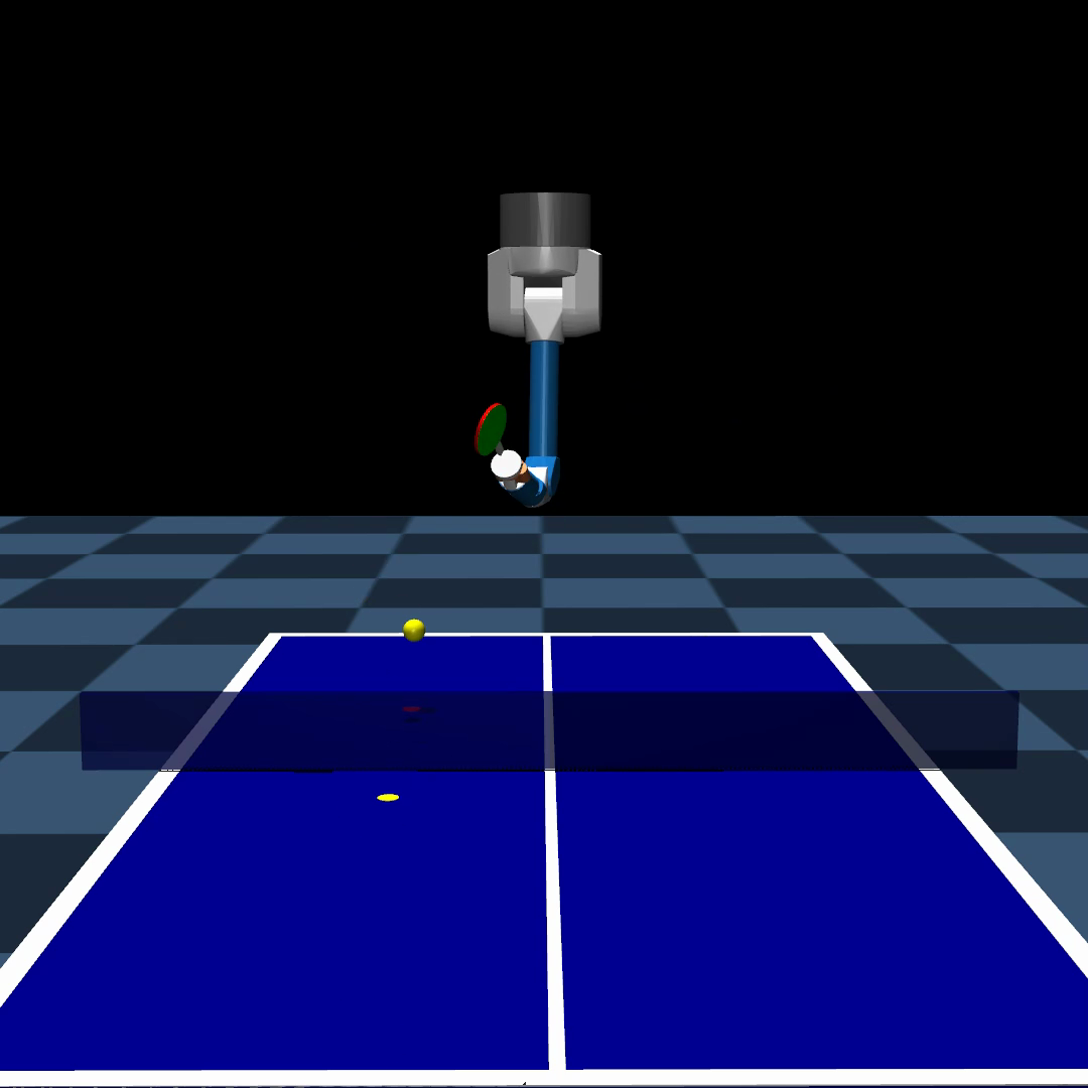}
	\end{minipage}\hfill
	\begin{minipage}[t!]{0.115\textwidth}
		\includegraphics[width=\textwidth]{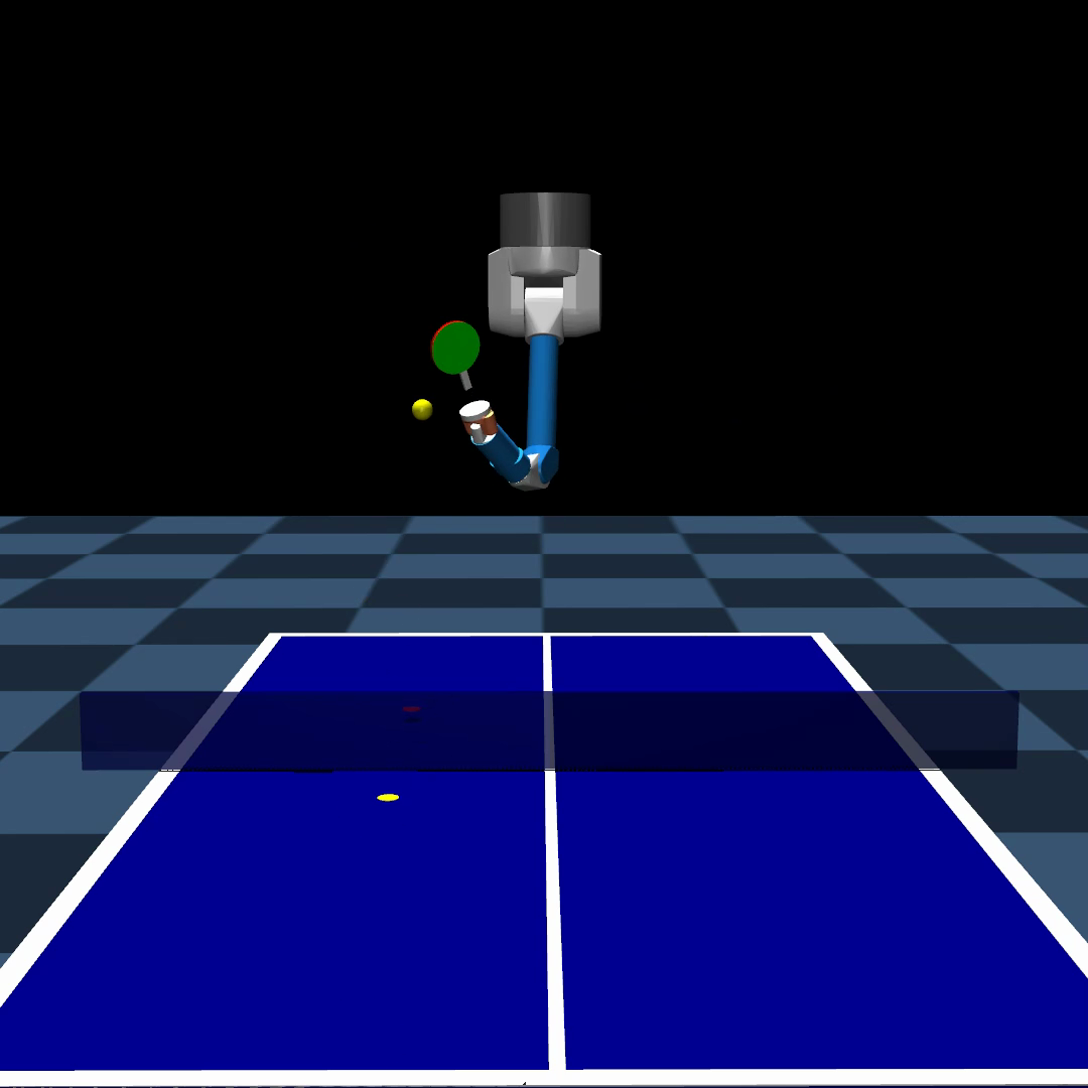}
	\end{minipage}\hfill
	\begin{minipage}[t!]{0.115\textwidth}
		\includegraphics[width=\textwidth]{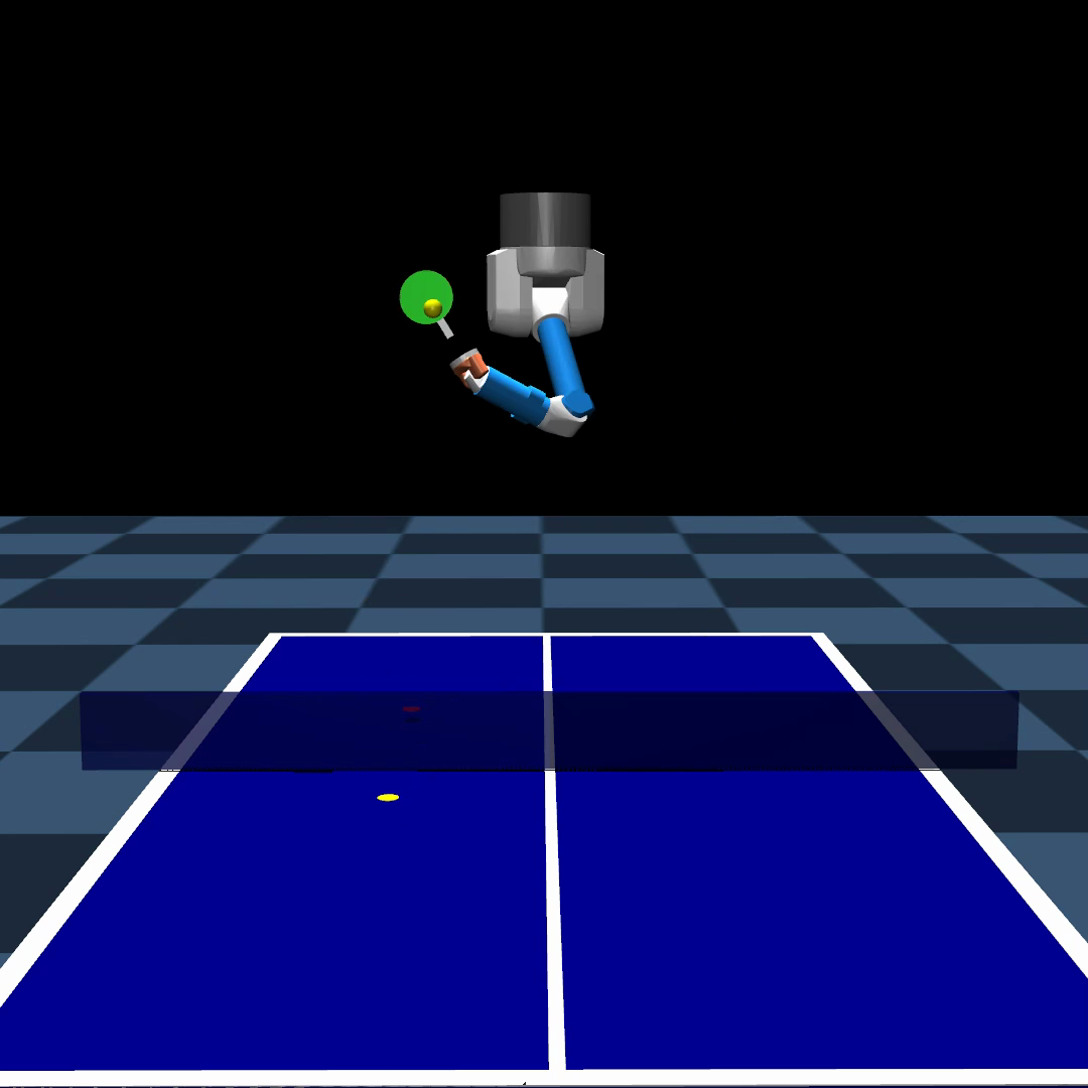}
	\end{minipage}\hfill
	\begin{minipage}[t!]{0.115\textwidth}
		\includegraphics[width=\textwidth]{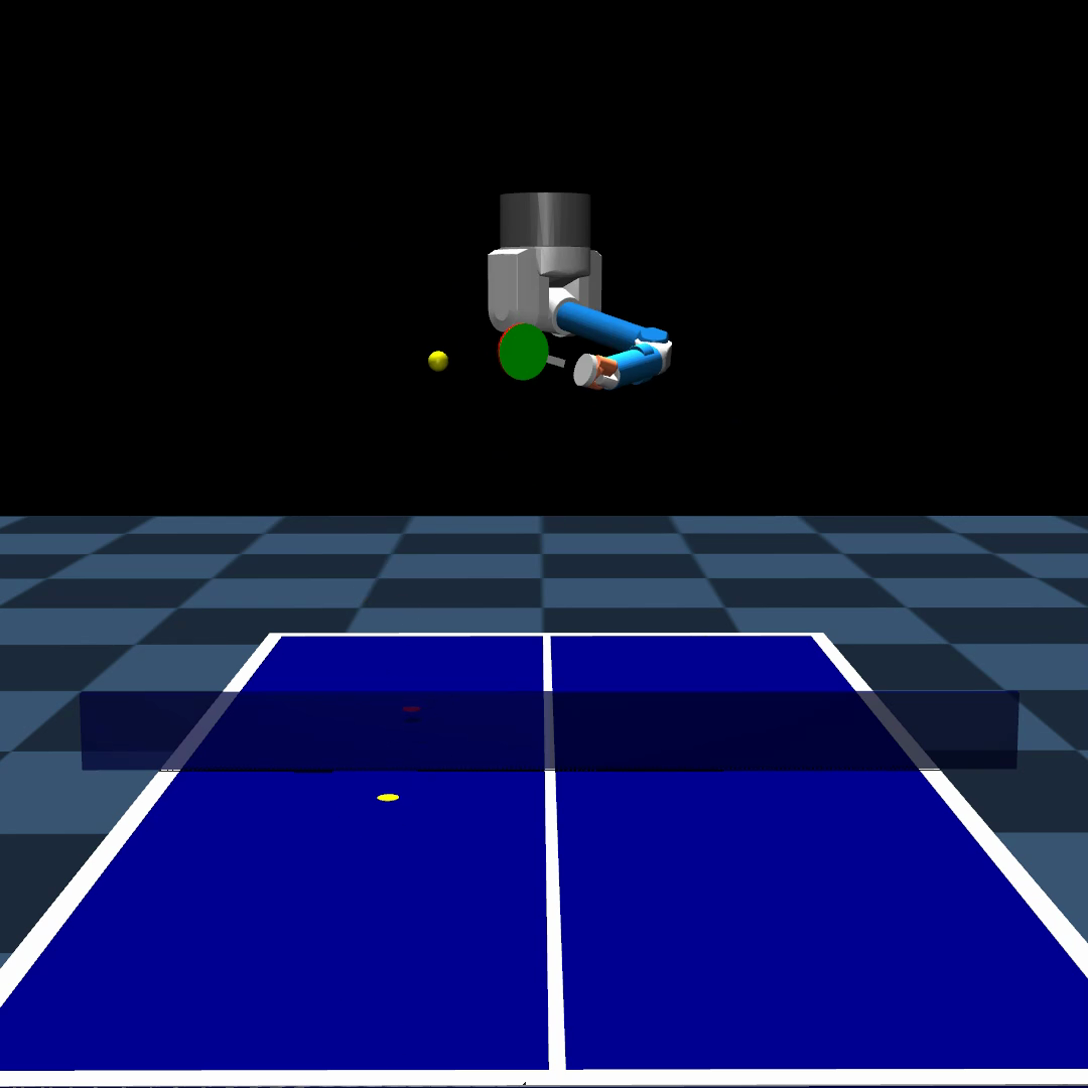}
	\end{minipage}\hfill
	\begin{minipage}[t!]{0.115\textwidth}
		\includegraphics[width=\textwidth]{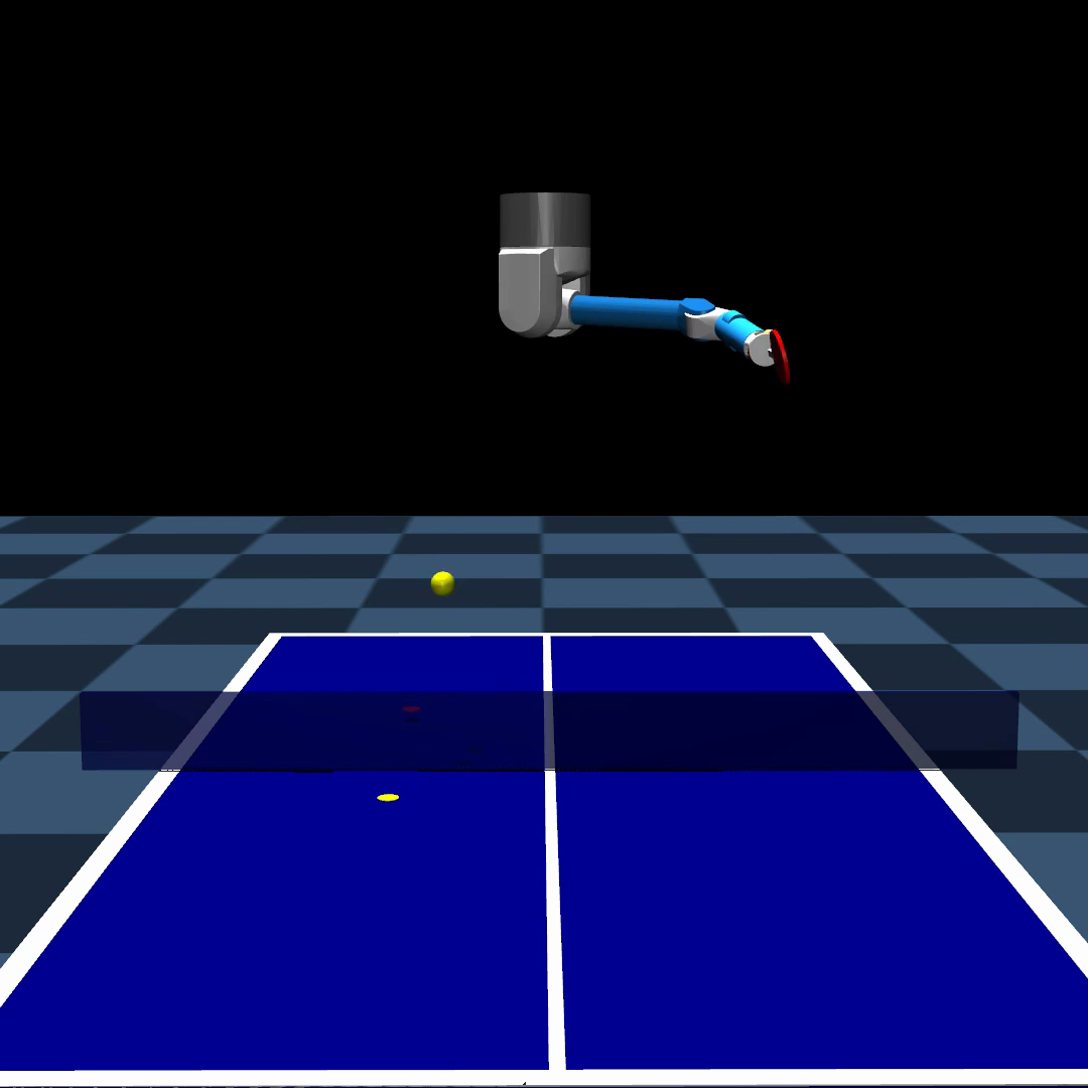}
	\end{minipage}\hfill
	\begin{minipage}[t!]{0.115\textwidth}
		\includegraphics[width=\textwidth]{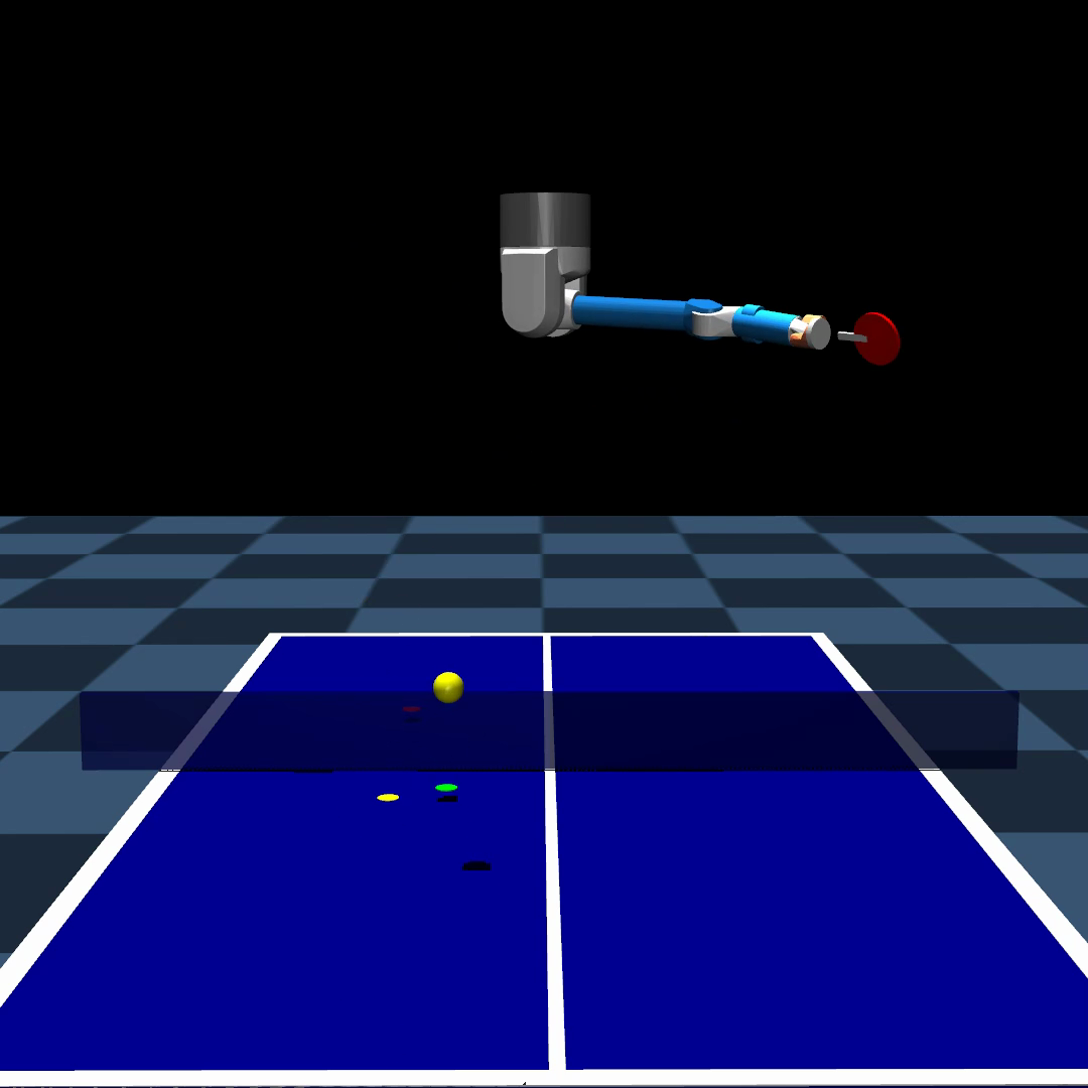}
	\end{minipage}\hfill
	\begin{minipage}[t!]{0.115\textwidth}
		\includegraphics[width=\textwidth]{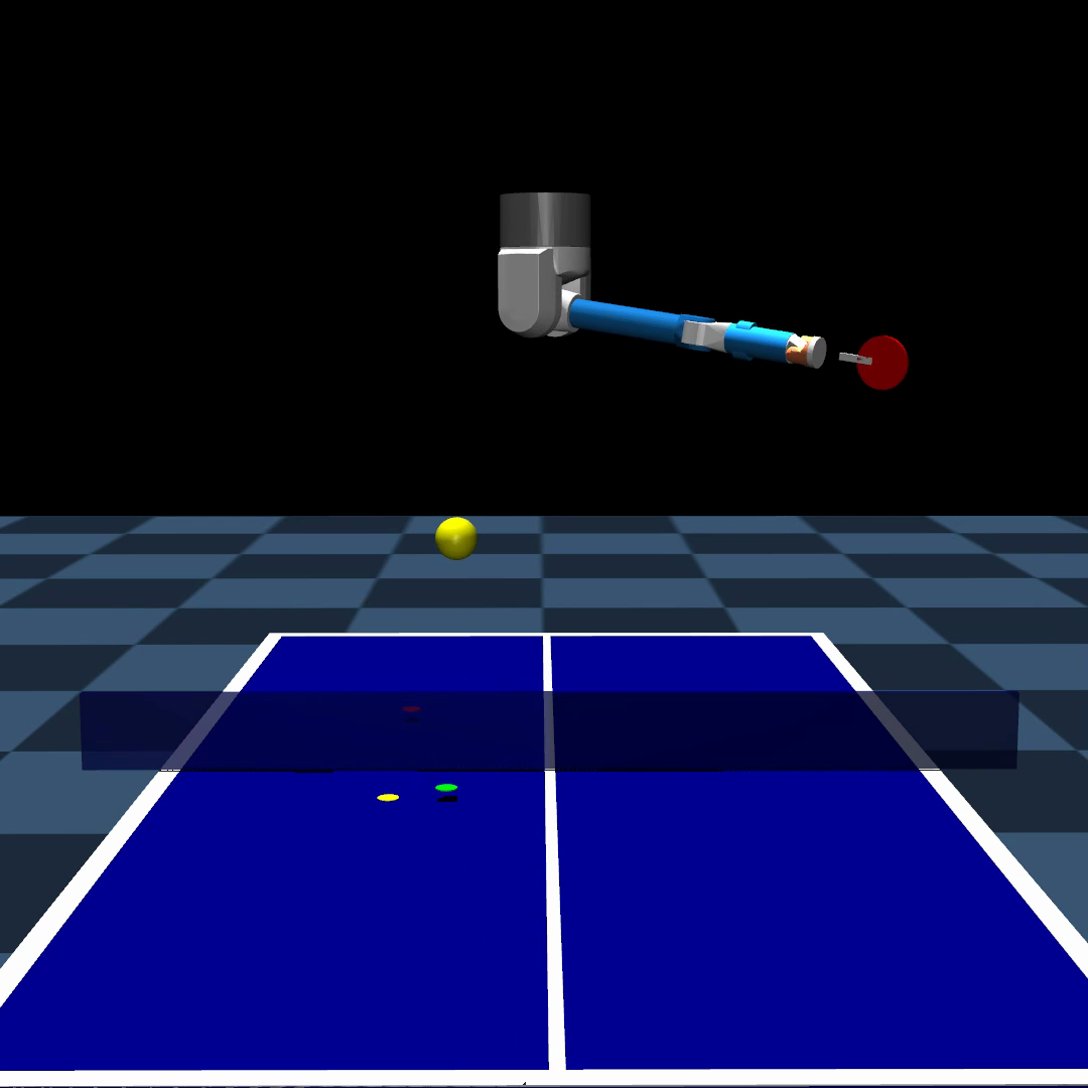}
	\end{minipage}\hfill
    \end{minipage}\hfill
	\begin{minipage}[t!]{\textwidth}
	    \vspace{2mm}
    	\begin{minipage}[t!]{0.115\textwidth}
    		\includegraphics[width=\textwidth]{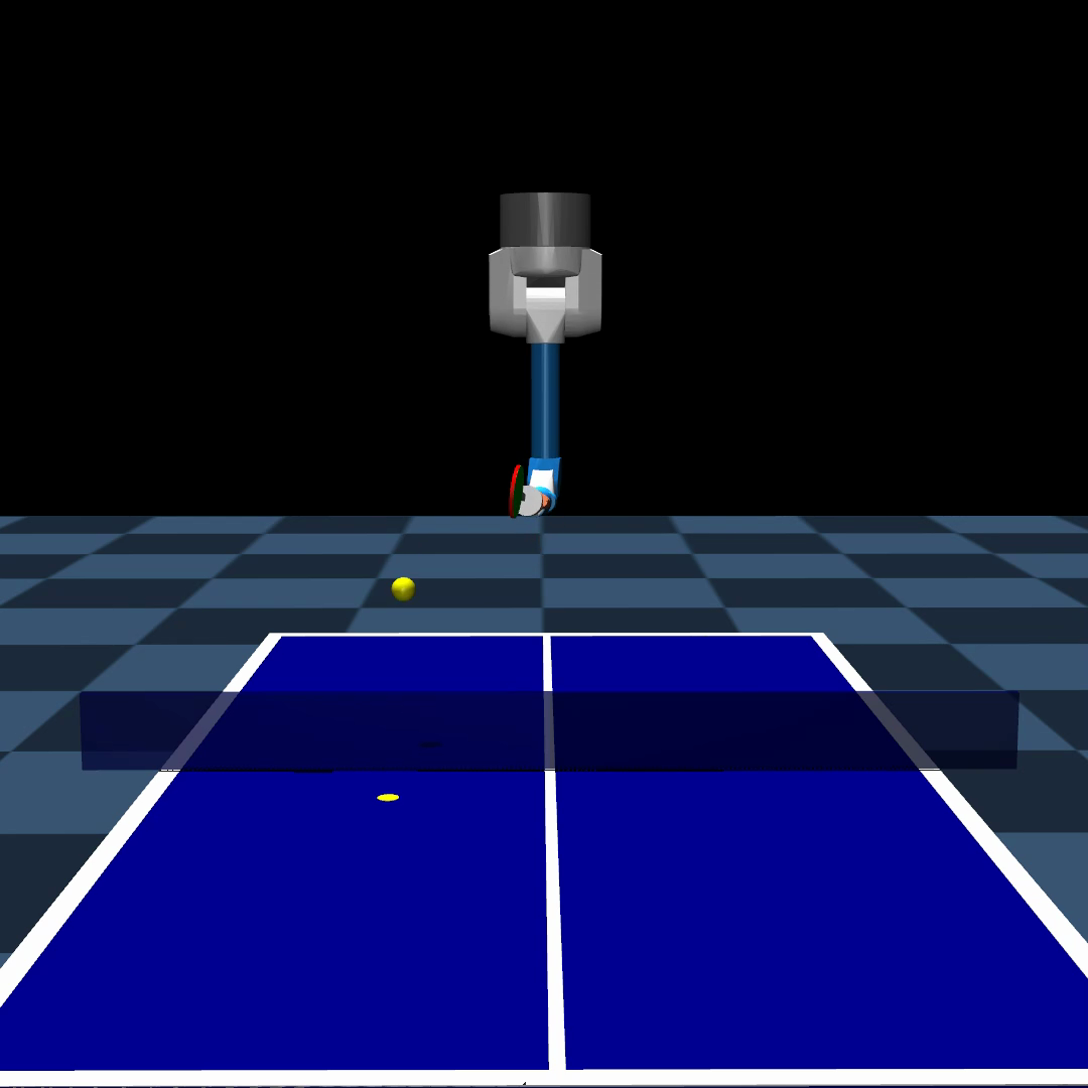}
    	\end{minipage}\hfill
    	\begin{minipage}[t!]{0.115\textwidth}
    		\includegraphics[width=\textwidth]{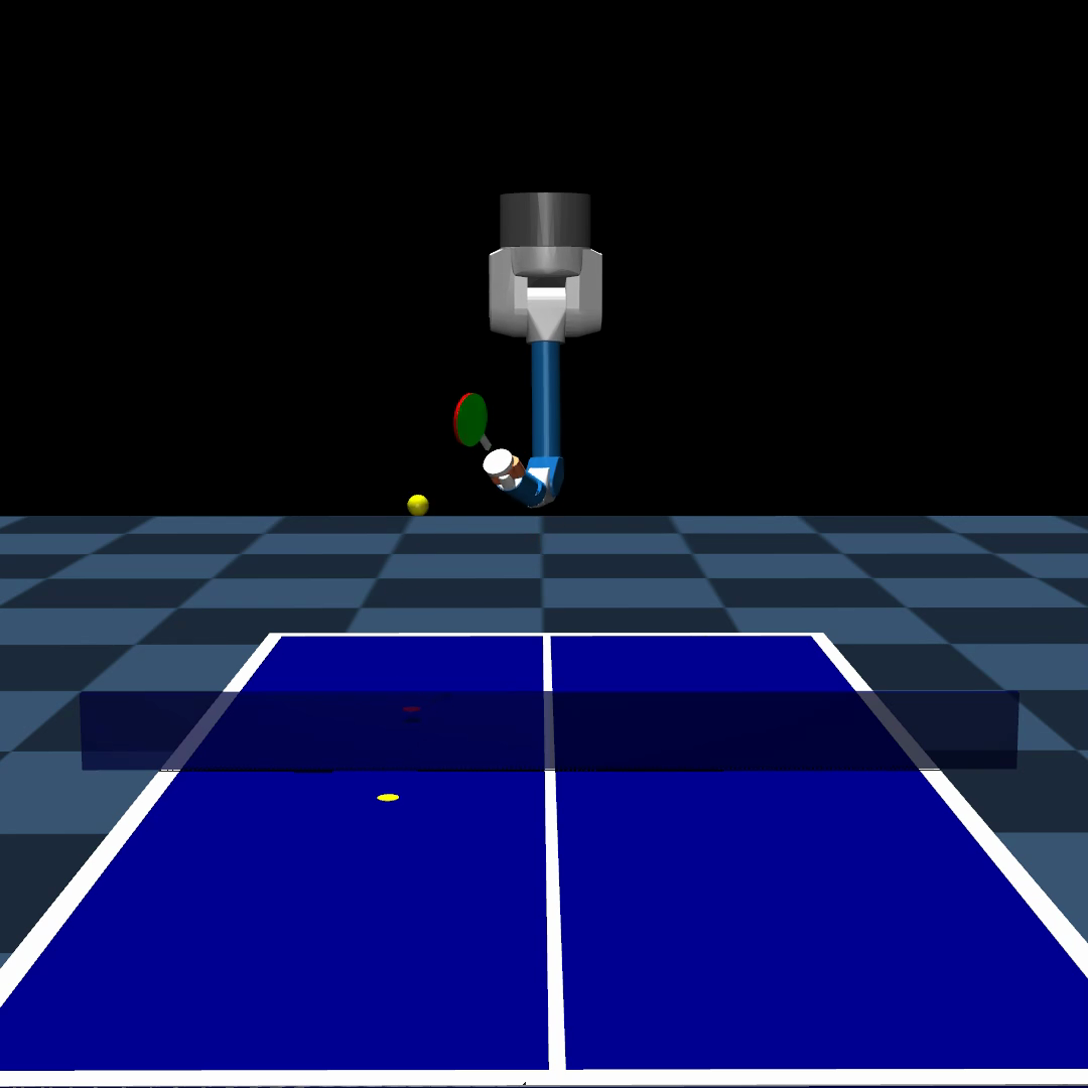}
    	\end{minipage}\hfill
    	\begin{minipage}[t!]{0.115\textwidth}
    		\includegraphics[width=\textwidth]{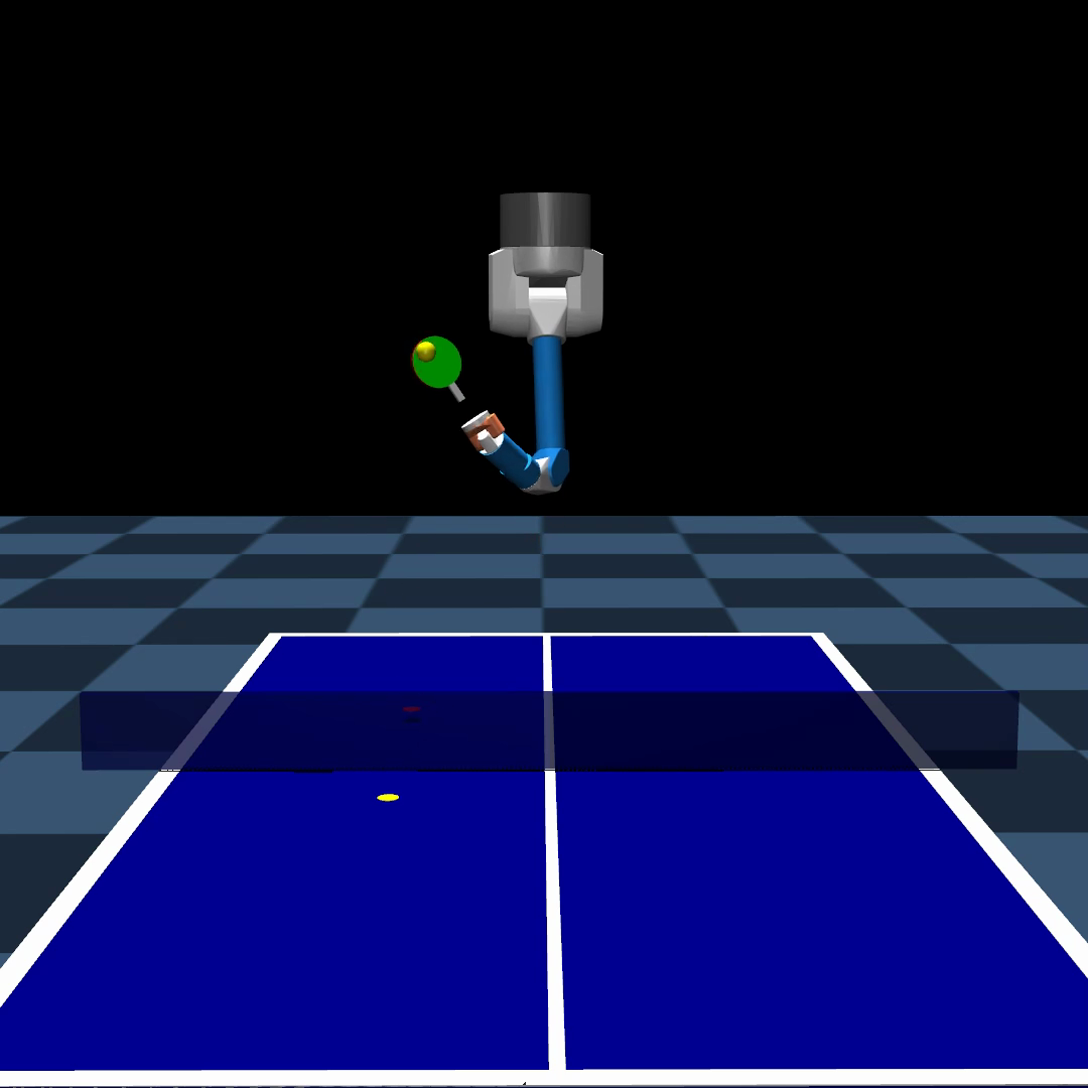}
    	\end{minipage}\hfill
    	\begin{minipage}[t!]{0.115\textwidth}
    		\includegraphics[width=\textwidth]{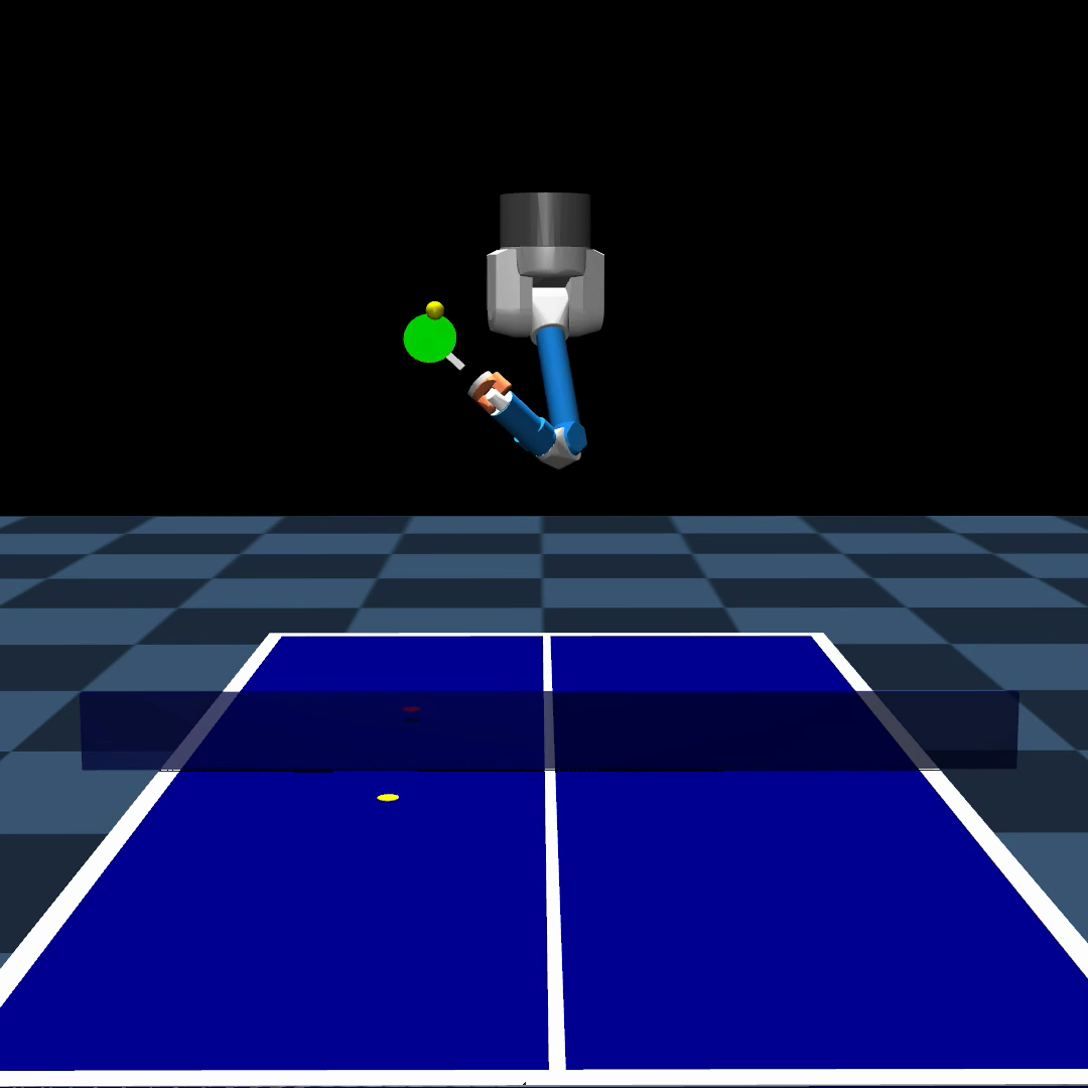}
    	\end{minipage}\hfill
    	\begin{minipage}[t!]{0.115\textwidth}
    		\includegraphics[width=\textwidth]{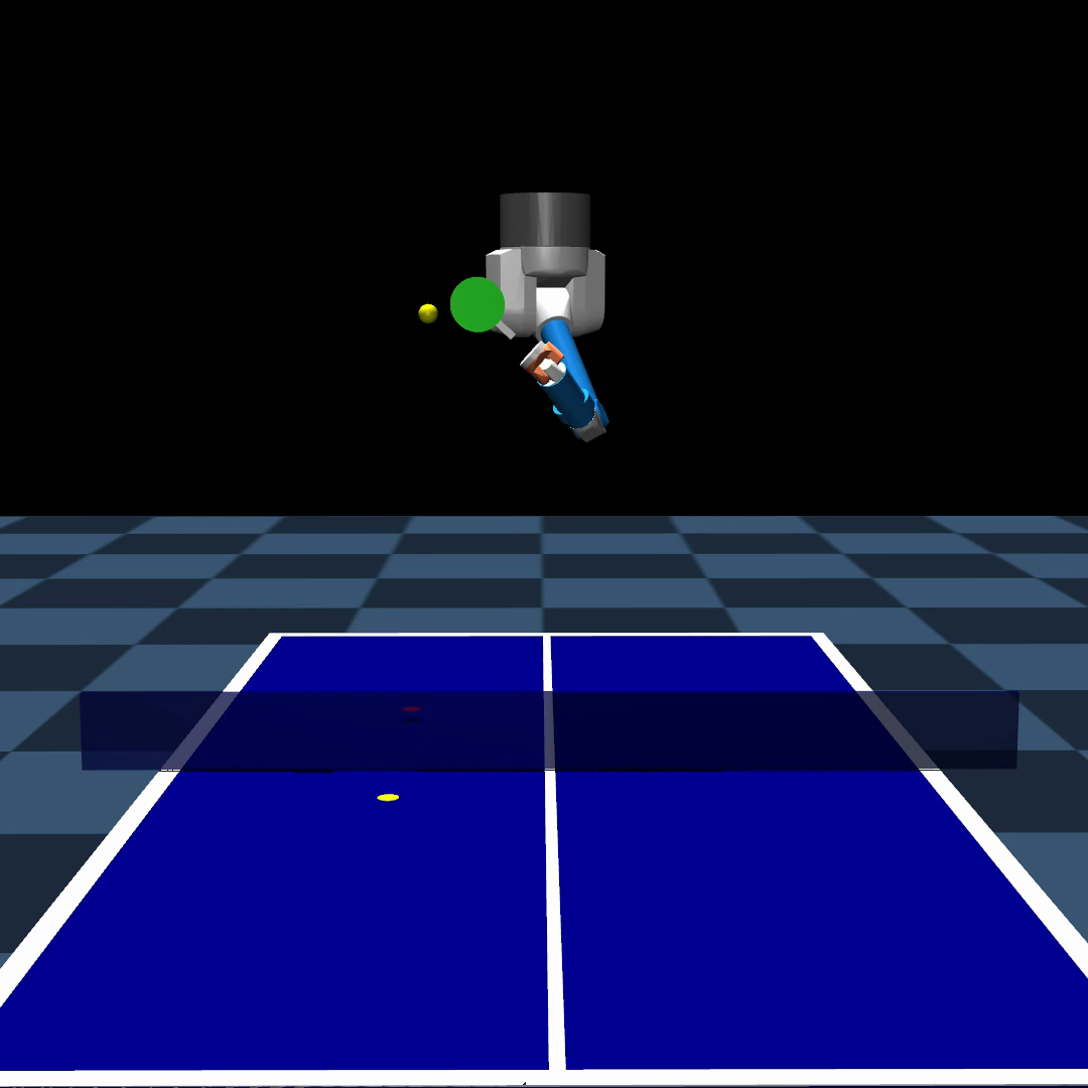}
    	\end{minipage}\hfill
    	\begin{minipage}[t!]{0.115\textwidth}
    		\includegraphics[width=\textwidth]{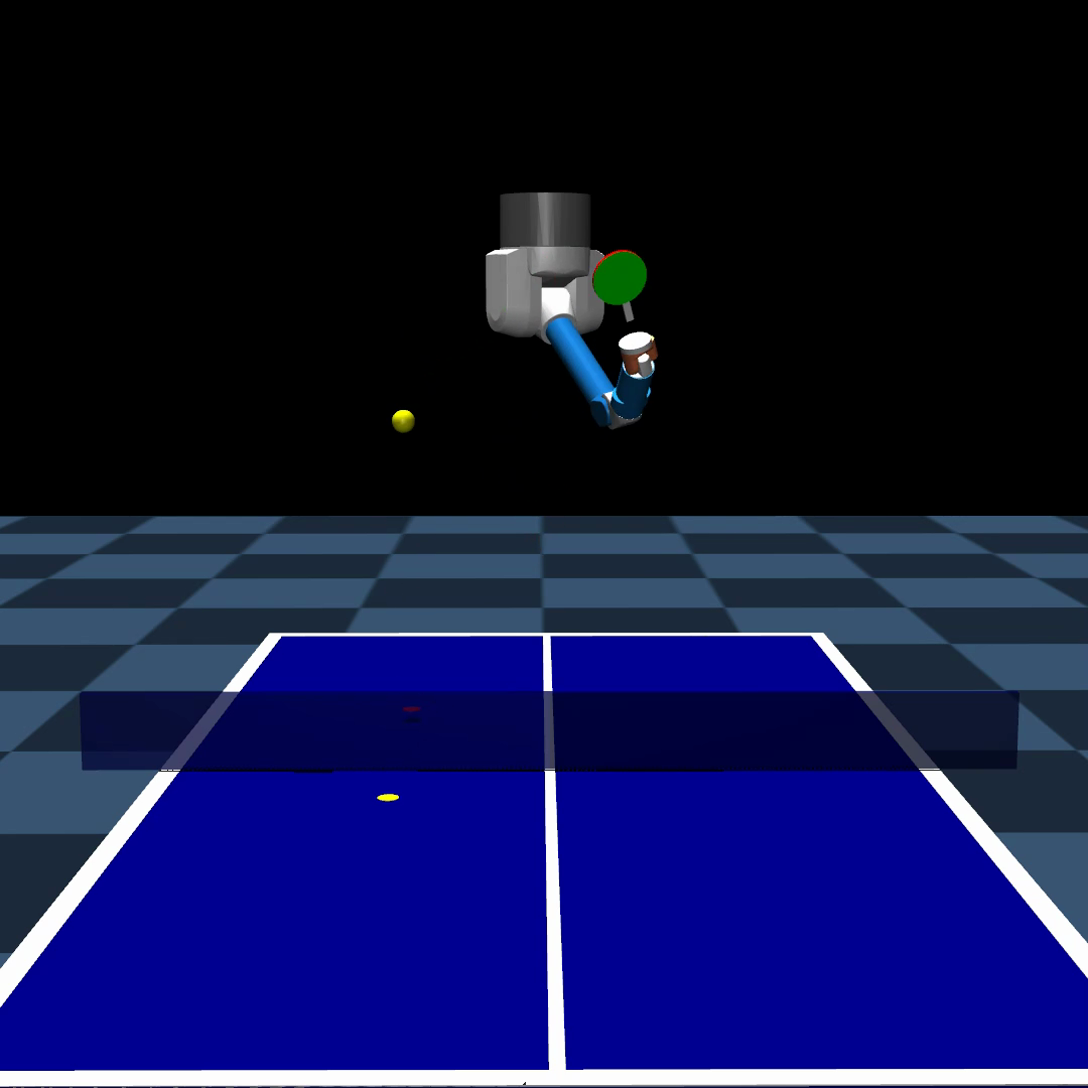}
    	\end{minipage}\hfill
    	\begin{minipage}[t!]{0.115\textwidth}
    		\includegraphics[width=\textwidth]{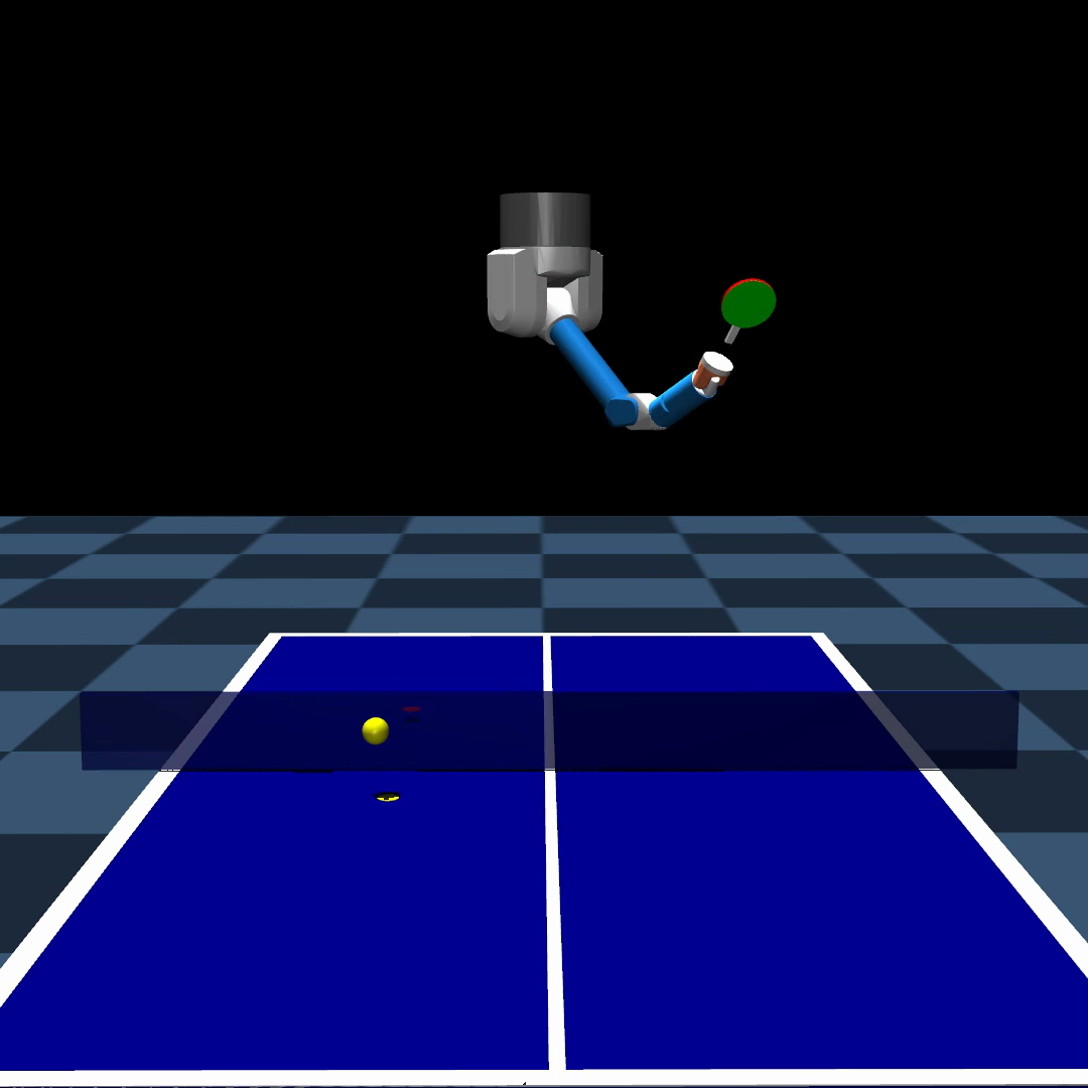}
    	\end{minipage}\hfill
    	\begin{minipage}[t!]{0.116\textwidth}
    		\includegraphics[width=\textwidth]{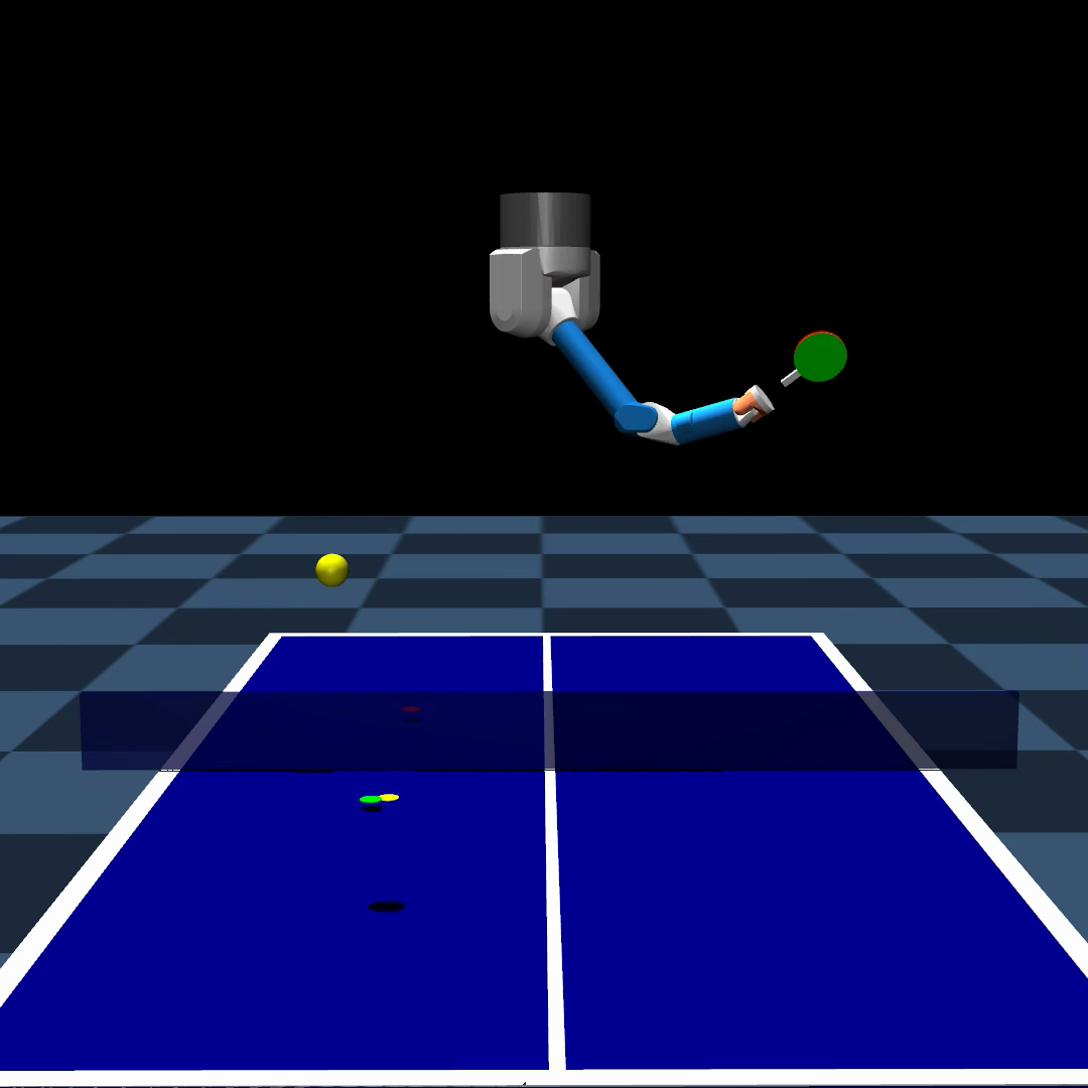}
    	\end{minipage}\hfill
	\end{minipage}\hfill
	\begin{minipage}[t!]{\textwidth}
	    \vspace{2mm}
    	\begin{minipage}[t!]{0.115\textwidth}
    		\includegraphics[width=\textwidth]{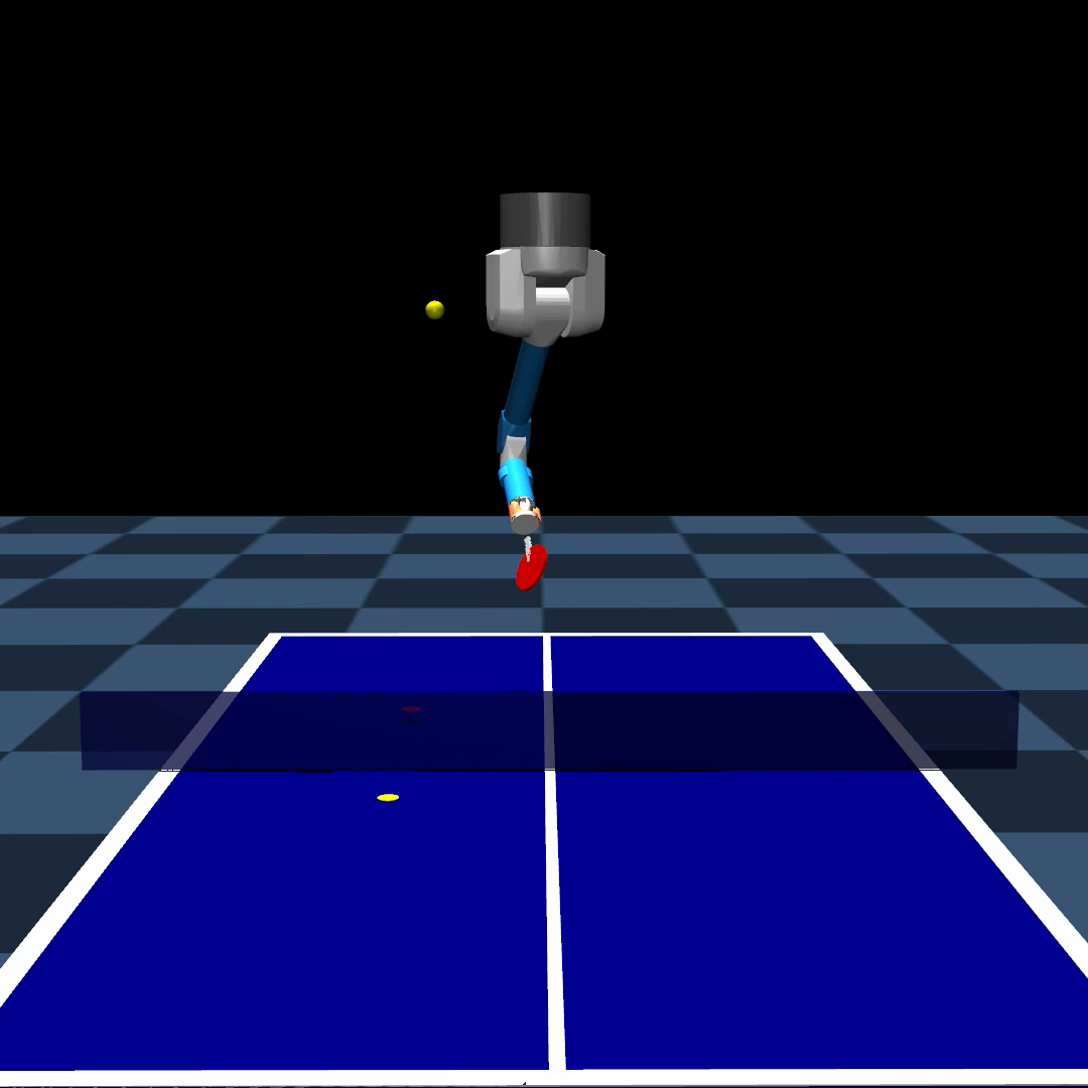}
    	\end{minipage}\hfill
    	\begin{minipage}[t!]{0.115\textwidth}
    		\includegraphics[width=\textwidth]{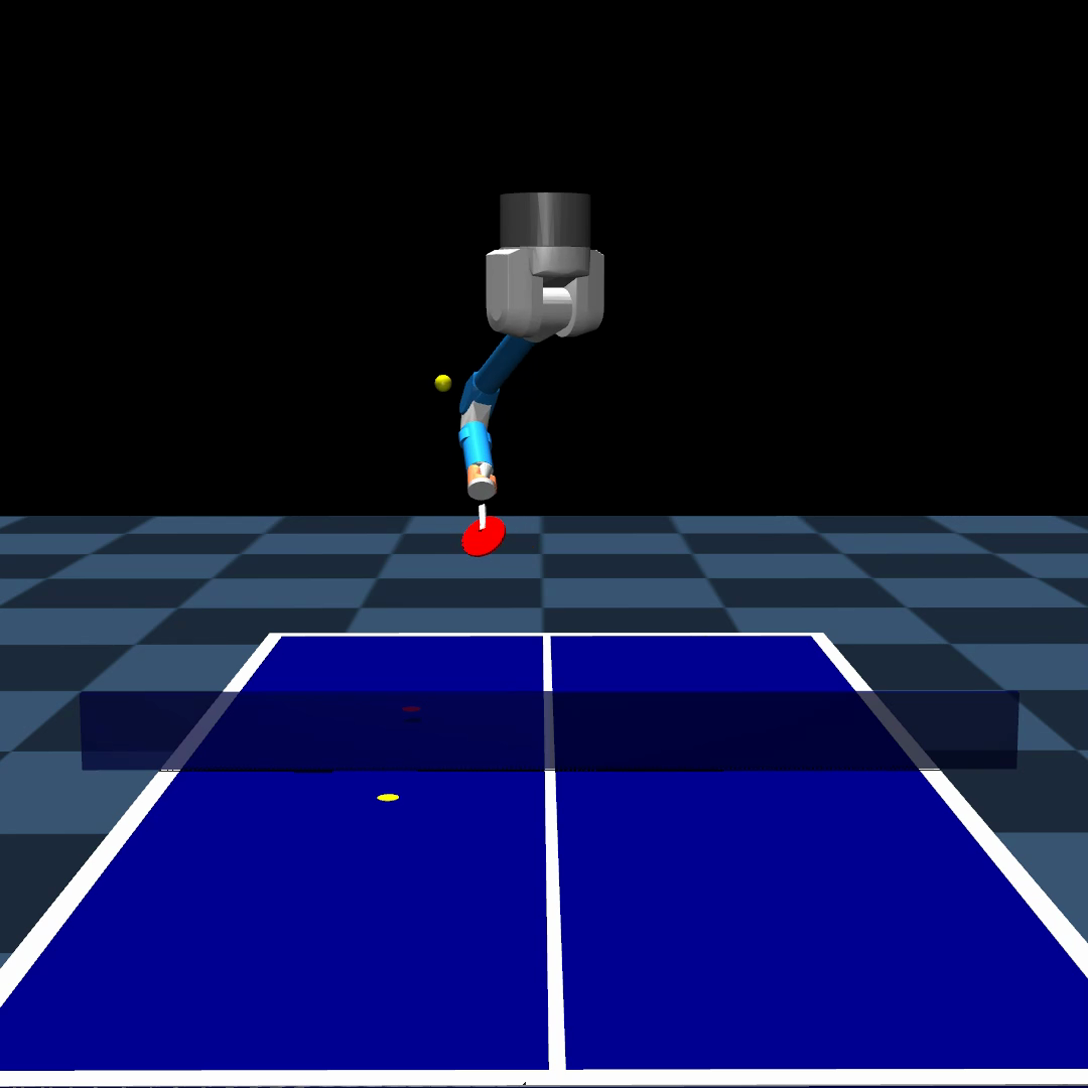}
    	\end{minipage}\hfill
    	\begin{minipage}[t!]{0.115\textwidth}
    		\includegraphics[width=\textwidth]{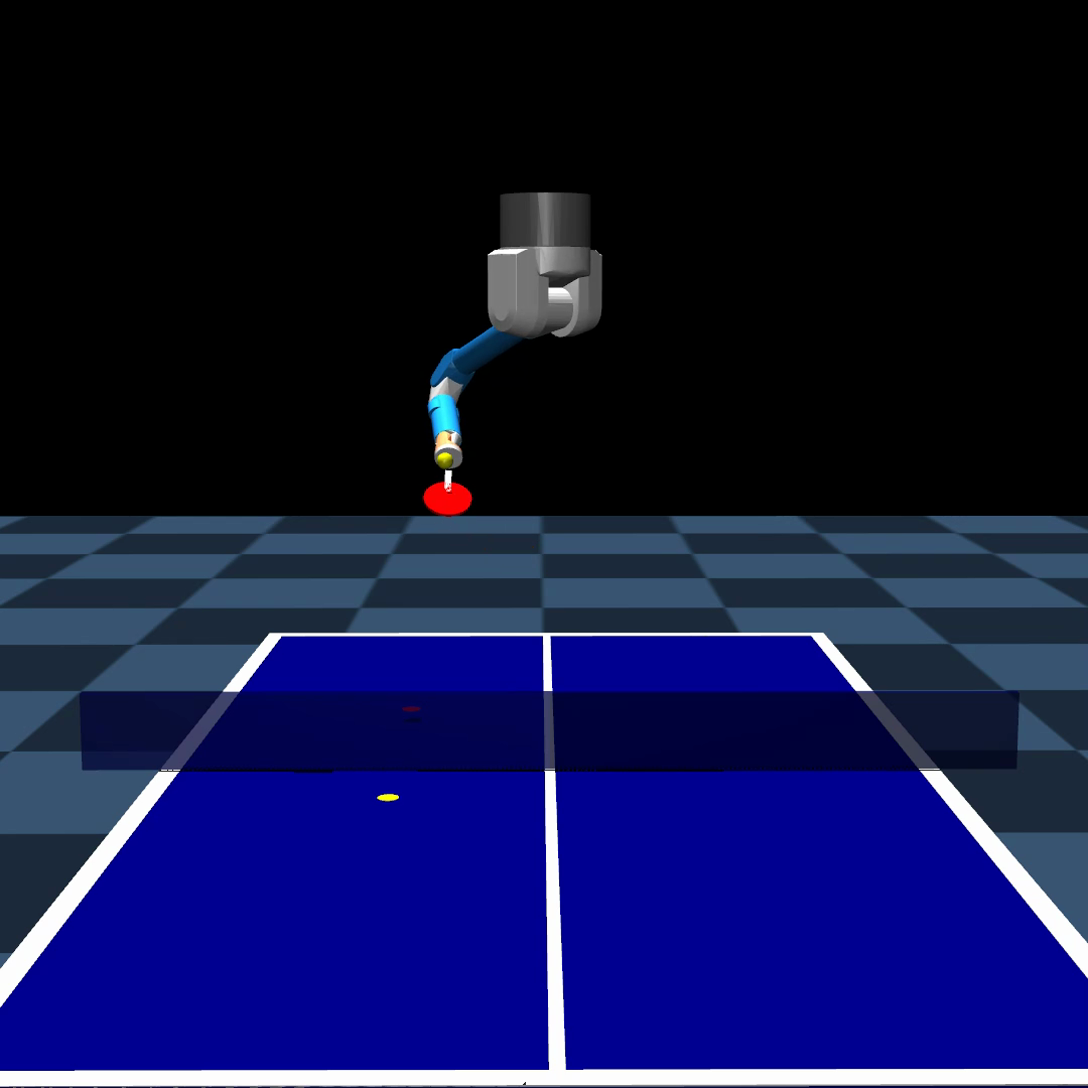}
    	\end{minipage}\hfill
    	\begin{minipage}[t!]{0.115\textwidth}
    		\includegraphics[width=\textwidth]{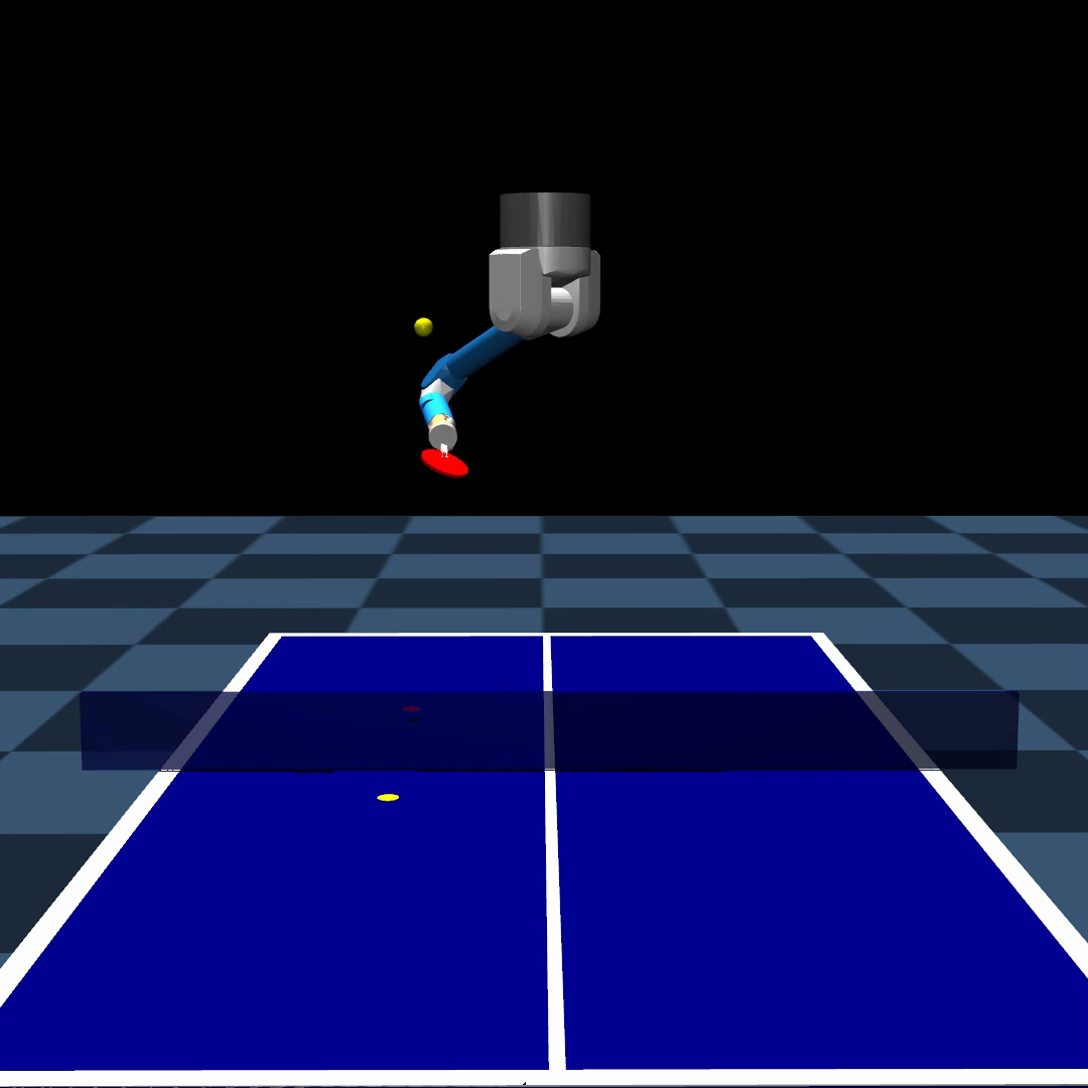}
    	\end{minipage}\hfill
    	\begin{minipage}[t!]{0.115\textwidth}
    		\includegraphics[width=\textwidth]{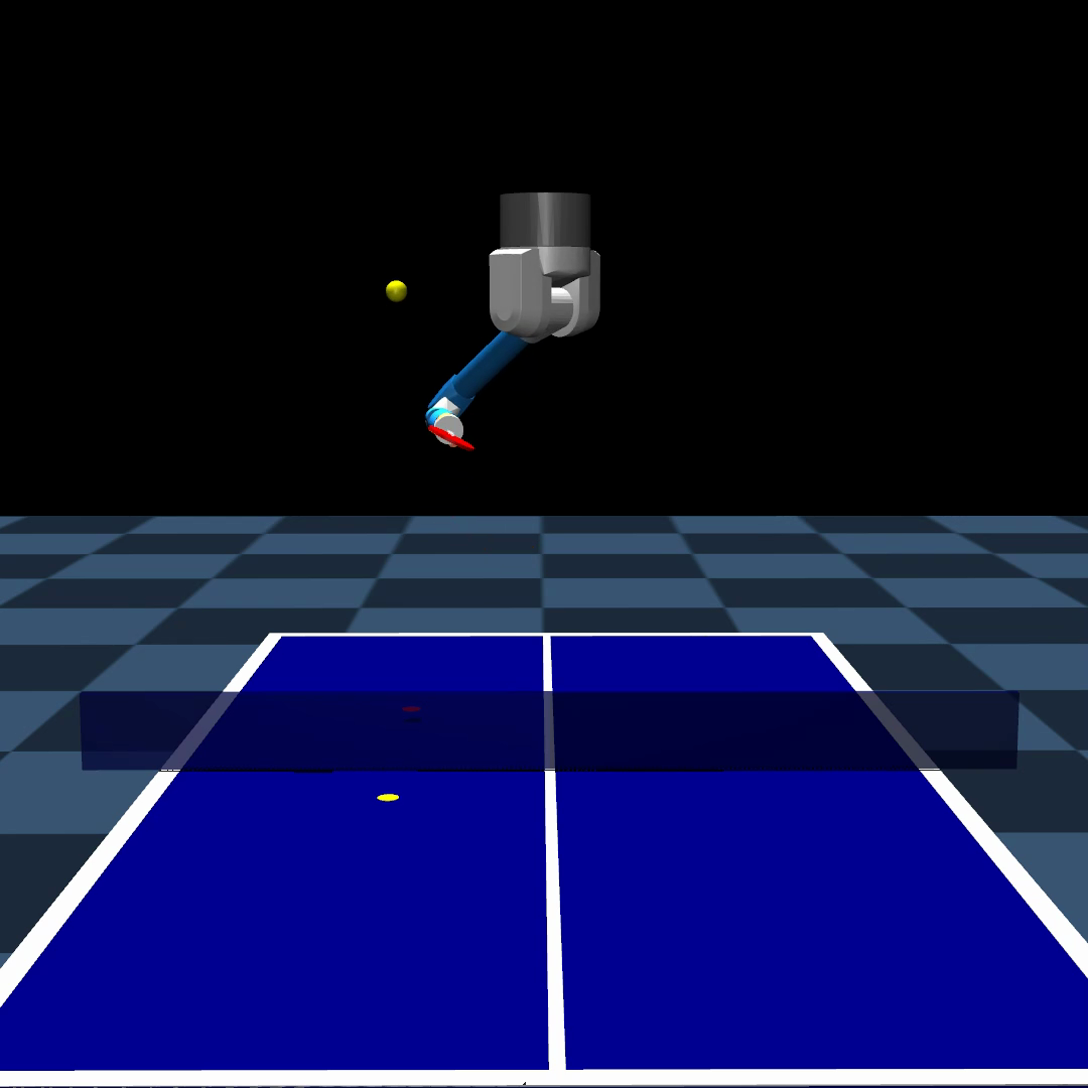}
    	\end{minipage}\hfill
    	\begin{minipage}[t!]{0.115\textwidth}
    		\includegraphics[width=\textwidth]{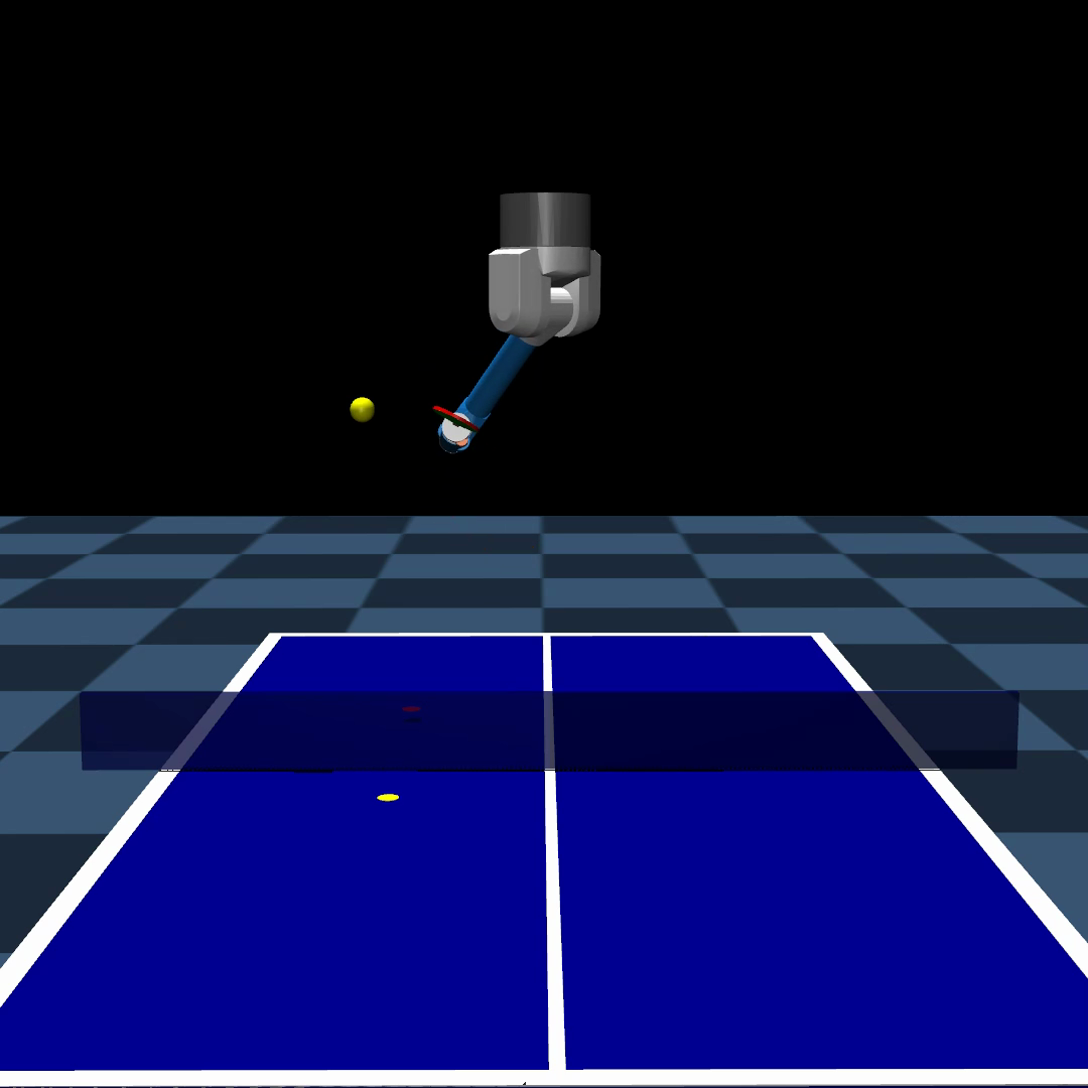}
    	\end{minipage}\hfill
    	\begin{minipage}[t!]{0.115\textwidth}
    		\includegraphics[width=\textwidth]{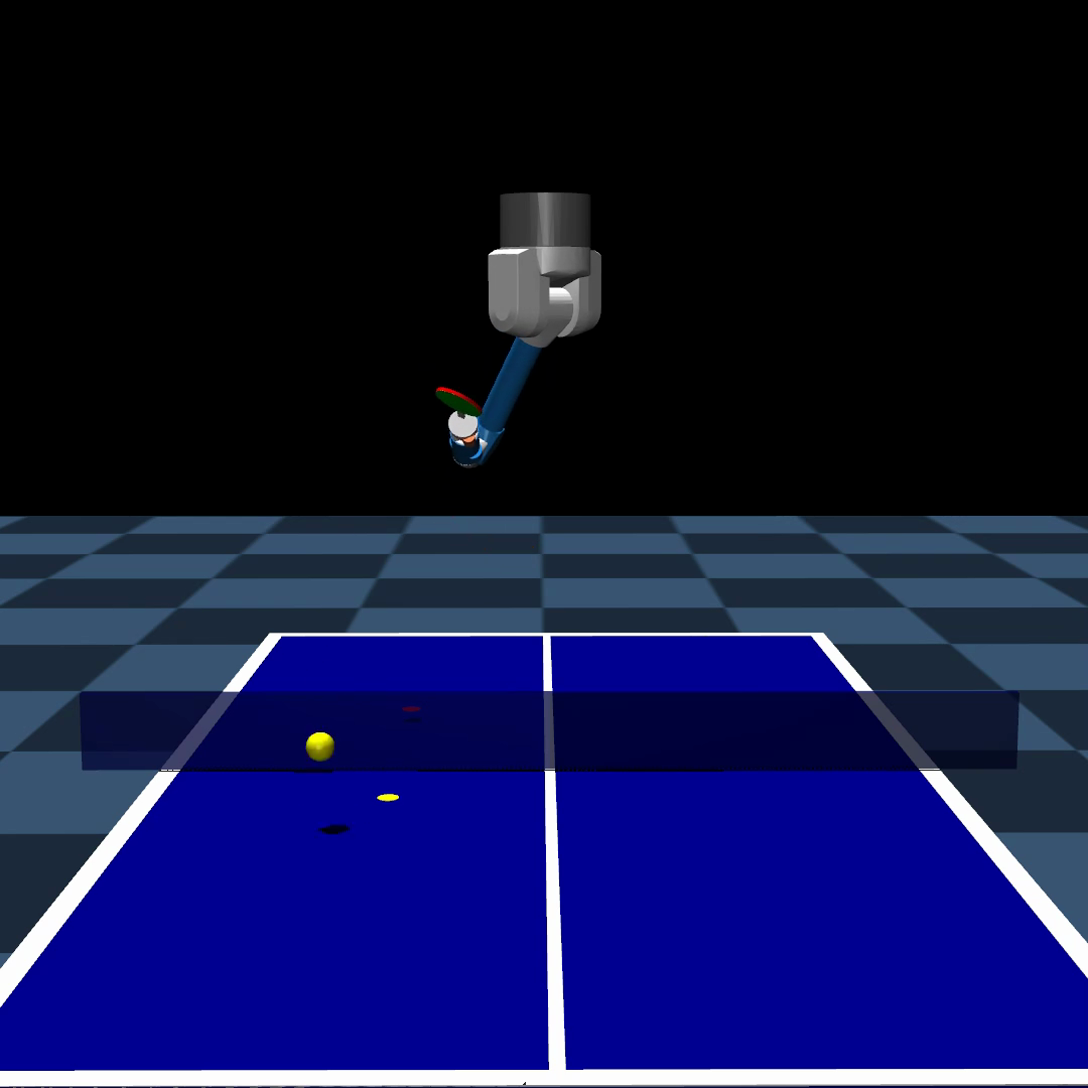}
    	\end{minipage}\hfill
    	\begin{minipage}[t!]{0.115\textwidth}
    		\includegraphics[width=\textwidth]{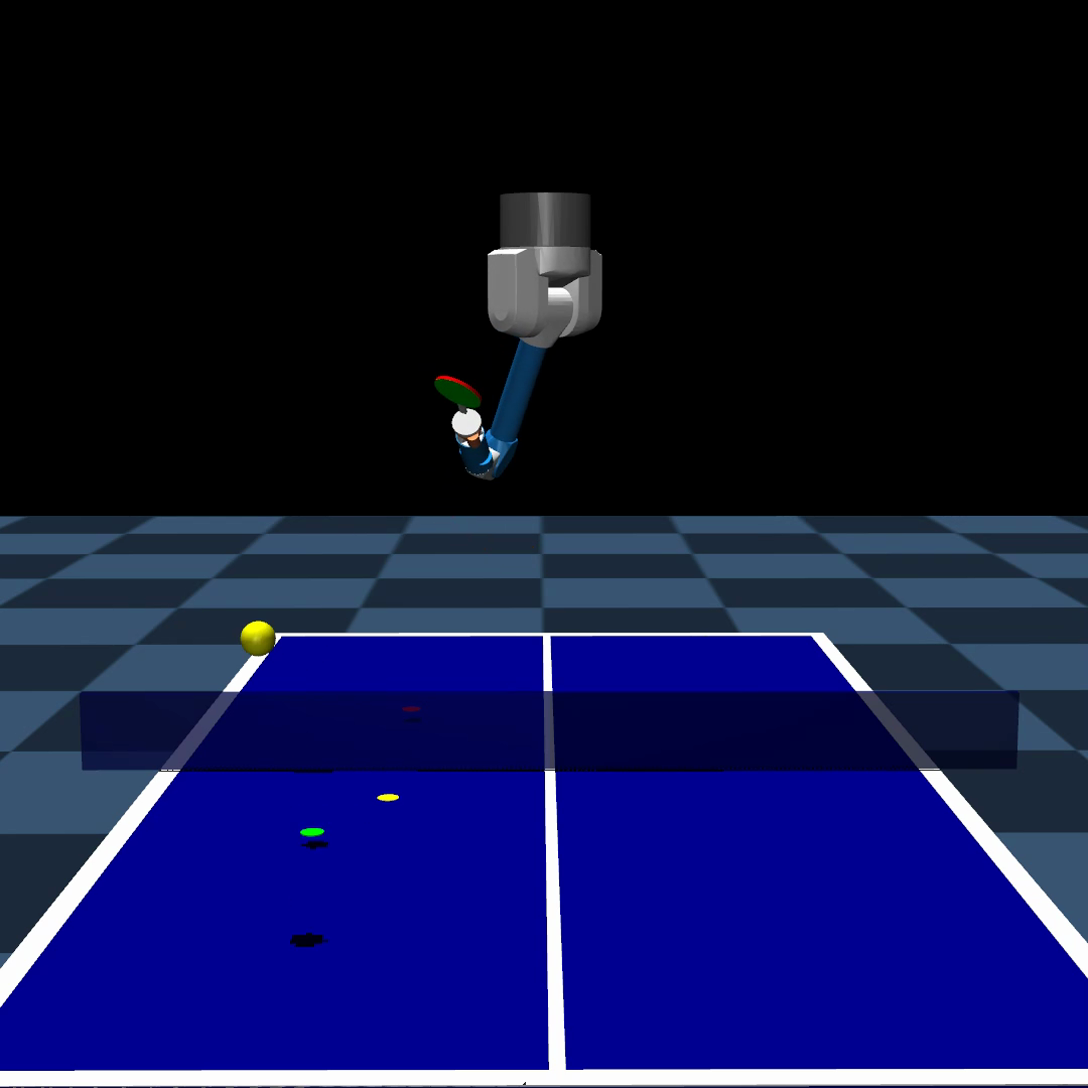}
    	\end{minipage}\hfill
	\end{minipage}\hfill
	\end{minipage}\hfill
	\vspace{-2mm}
   \caption{\textbf{Versatile Strikes for the Table Tennis (TT) Experiment} illustrated for a fixed context. The robot can hit the (yellow) ball in various ways, also indicated through the different colors of the racket sides. Note that the red and yellow dots on the table are markers for the serving and desired landing position respectively.}
   \label{fig::exps_tt_episodic::versatile_strikes}
   \vspace{-1.5mm}
\end{figure}

{\bf Table Tennis.} The task of the robot is to return different incoming balls to desired targets on the opponent's table side in different ways. We consider a four dimensional context space, including ball's initial serve position and ball's target landing position (right and left half of table Fig. \ref{fig::exps:tt_episodic::ctxt_distr})). For both parts of the context, we fix the z position and vary the x and y coordinates.
We incrementally add 50 components in our method, while HiREPS and LaDiPS directly start with 50 components. We have run all algorithms over 20 seeds/trials and report the performance in Fig. \ref{fig::exps_tt_episodic::rewards})\footref{exclude_last_comp}, in which the mean reward with two times the standard error is plotted. The component-wise context distributions are spread among the context space (Fig. \ref{fig::exps:tt_episodic::ctxt_distr})) and allow each component to locally specialize on a context region. This high coverage of the context space allows for a high reward achievement, outperforming HiREPS and LaDiPS.
In Fig. \ref{fig::exps_tt_episodic::versatile_strikes}), we show three different striking skills sampled from our trained MoE model, for a fixed context, i.e., fixed serving and desired ball position. The first two skills use the green side of the racket to hit the ball (forehand), while the third skill uses the red side of the racket (backhand) to hit the ball. In contrast to the second strike, the first one performs a smash-like strike and ends it with the red side of the racket pointing to the camera. 
\section{Conclusion}\label{sec::conclusion}
We proposed a new objective for learning contextual and versatile Mixtures of Experts (MoE) models. 
We based our objective on a maximum entropy formulation to increase the versatility of the solutions and introduced a curriculum to allow the components to specialize. 
Our formulation also allows for easy online adaptation of the model complexity during training. 
We conducted an ablation to show the importance of the individual parts of our objective. 
Further, we showed that our method learns precise and versatile solutions and outperforms the baseline on sophisticated simulated robotic tasks. 
This work aims at presenting a mathematically well-founded method and demonstrates its general feasibility on various challenging tasks. 
Currently, the major drawback of our approach is sample efficiency, as we do not share experience between the components. 
We intend to address this issue in future work, e.g., by intra-option learning.
Another direction for future research is extending the approach to more complex models, such as non-linear mixtures of experts. We expect to need fewer components to cover the whole context space with more complex component model representations.



\clearpage
\acknowledgments{Calculations for this research were conducted on the high performance computer of the state of Baden-Württemberg.}


\bibliography{ref}  
\newpage
\appendix
\section{Derivations}\label{sec::app::derivations}
In the following we derive our objectives from Section \ref{sec::SpecializingVersatileMixtureOfExpertModels}. 
\subsection{Maximum Entropy Skill Learning with Curriculum}
We start our derivation from the KL-regularized maximum entropy objective
\begin{align}\label{eq::appendix::maxEntrKLOBJ}
    \max_{\joint{\pi}} \mathbb{E}_{\pi(\cvec c)}\left[\mathbb{E}_{\policy} \left[\reward\right] + \alpha \textrm{H}\left[\policy\right]\right] - \beta\KL{\statedistr}{p(\cvec c)}
\end{align}
We use following identities
\begin{align*}
    \policy =& \sum_o \frac{\statecomp \mgating}{\statedistr}\comppi\\
    \log \policy &= \log \frac{\comppi \gatingpi}{\responsibilitypi} \\
    \statedistr &= \sum_o \statecomp \mgating\\
    \log \statedistr &=\log \frac{\statecomp \mgating}{\gatingpi}
\end{align*}
and insert them into our objective in Equation (\ref{eq::appendix::maxEntrKLOBJ}), which leads to 
\begin{align}\label{apdxEnrolledObj}
    L &= \sum_o \mgating \mathbb{E}_{\statecomp}\left[\mathbb{E}_{\comppi} \left[\reward -\alpha \log \comppi 
      -\alpha \log \gatingpi  + \alpha \log \responsibilitypi \right]\right] \nonumber \\
    &+\sum_o \mgating \mathbb{E}_{\statecomp}\left[-\beta \log \statecomp  +\beta \log \gatingpi -\beta \log \mgating \right] . 
\end{align}
Note that we have dropped the $\log p(\cvec c)$ term since we assume $p(\cvec c)$ to be uniformly distributed in a given interval.\\
By further rearranging the terms we can reformulate this objective as 
\begin{align}
    L &= \sum_o \mgating \mathbb{E}_{\statecomp} \left[ \mathbb{E}_{\comppi}\left[\reward -\alpha \log \comppi + \alpha \log \responsibilitypi \right]\right] \\ \nonumber
    &+\sum_o \mgating \mathbb{E}_{\statecomp} \left[-\beta \log \statecomp +(\beta-\alpha) \log \gatingpi -\beta \log \mgating \right]. 
\end{align}
As stated in the paper, we can not optimize this objective. Therefore we introduce the auxiliary distributions $\tilderesppi$ and $\tildegating$ as
\begin{align*}
    L &= \sum_o \mgating \mathbb{E}_{\statecomp} \left[\mathbb{E}_{\comppi} \left[\reward -\alpha \log \comppi + \alpha \log \responsibilitypi +\alpha \log \tilderesppi \right. \right. \\
    & \left. \left. -\alpha \log \tilderesppi \right]\right] + \sum_o \mgating \mathbb{E}_{\statecomp} \left[-\beta \log \statecomp  -\beta \log \mgating +(\beta-\alpha) \log \gatingpi \right. \\
    & \left. +(\beta-\alpha)\log \tildegating -(\beta - \alpha) \tildegating\right]. 
\end{align*}
We can rearrange the terms further to 
\begin{align}\label{eq::appendix::DecomposedObj}
    L &= \sum_o \mgating \mathbb{E}_{\statecomp} \left[ \mathbb{E}_{\comppi} \left[\reward +\alpha \log \tilderesppi\right] +(\beta-\alpha)\log \tildegating \right] \nonumber \\
    &+\alpha \mathbb{E}_{\mgating, \comppi}\left[\mathcal{H}\left(\comppi\right)\right] +\beta \mathbb{E}_{\mgating}\left[\mathcal{H}\left(\statecomp\right)\right] +\beta \mathcal{H}\left(\mgating\right) \nonumber \\ &
      +\alpha \mathbb{E}_{\statedistr, \policy}\left[\textrm{KL}\left(\responsibilitypi||\tilderesppi\right)\right] +(\beta - \alpha) \mathbb{E}_{\statedistr}\left[\textrm{KL}\left(\gatingpi || \tildegating \right)\right].
\end{align}

\subsection{Lower-Bound Decomposition for Component Distributions}
By observing that not all terms depend on $\comppi$, we extract only the important ones from the objective in Equation (\ref{eq::appendix::DecomposedObj}) as
\begin{align*}
    \Tilde{L}_c &= \sum_o \mgating \mathbb{E}_{\statecomp} \left[\mathbb{E}_{\comppi} \left[\reward  +\alpha \log \tilderesppi\right]\right] +\alpha \mathbb{E}_{\mgating, \comppi}\left[\mathcal{H}\left(\comppi\right)\right] \nonumber \\
    &+\alpha \mathbb{E}_{\statedistr, \policy}\left[\textrm{KL}\left(\responsibilitypi||\tilderesppi\right)\right].
\end{align*}
Since the KL is always positive, we can write the lower bound to this objective as 
\begin{align*}
    L_c &= \sum_o \mgating \mathbb{E}_{\statecomp} \left[\mathbb{E}_{\comppi} \left[\reward +\alpha \log \tilderesppi\right]\right] +\alpha \mathbb{E}_{\mgating, \comppi}\left[\mathcal{H}\left(\comppi\right)\right] \nonumber, 
\end{align*}
such that $\Tilde{L}_c \geq L_c$ always holds. After maximizing $L_c$ w.r.t. $\comppi$ we tighten the lower bound by updating the responsibilities as described in the paper.

\subsection{Lower Bound Decomposition for Component-wise Context Distributions}

By neglecting all terms which do not depend on $\statecomp$ in objective (\ref{eq::appendix::DecomposedObj}), we can write the objective for optimizing w.r.t. $\statecomp$ as 
\begin{align*}
    \Tilde{L}_{k} &= \sum_o \mgating \mathbb{E}_{\statecomp} \left[ L_c(o, \cvec c) +(\beta-\alpha)\log \tildegating \right]+\beta \mathbb{E}_{\mgating}\left[\mathcal{H}\left(\statecomp\right)\right] \nonumber \\
    & + (\beta - \alpha) \mathbb{E}_{\statedistr}\left[\textrm{KL}\left(\gatingpi || \tildegating \right)\right], \nonumber
\end{align*}
where 
\begin{align*}
    L_c(o, \cvec c) =& \mathbb{E}_{\comppi} \left[\reward  +\alpha \log \tilderesppi \right] + \alpha \mathcal{H}(\comppi).
\end{align*}
Again we can observe that the KL term is always positive and we can write the lower bound to $\Tilde{L}_{k}$ as
\begin{align*}
    L_{c, k} &= \sum_o \mgating \mathbb{E}_{\statecomp}\left[L_c(o,\cvec s) + (\beta - \alpha) \log \tildegating\right]+ \beta \mathbb{E}_{\mgating}\left[\mathcal{H}\left(\statecomp\right)\right],
\end{align*}
such that $ \Tilde{L_{k}}\geq L_{c,k}$ holds. By setting the auxillary distributions to the responsibilities of the updated model, we tighten the bound.

\subsection{Lower Bound Decomposition for Prior Weights}

Finally we can write down the objective for updating the prior weights $\mgating$ as 
\begin{align}\label{eq::appendix::WeightUpdate}
    L_p &= \sum_o L_{c,k}(o) + \beta \mathcal{H}\left(\mgating\right),
\end{align}
where
\begin{align*}
    L_{c,k}(o) =& \mgating \mathbb{E}_{\statecomp} \left[L_c(o,\cvec s) + (\beta - \alpha) \log \tildegating\right] + \beta \textrm{H}\left(\statecomp\right).
\end{align*}

Some context regions naturally lead to low reward due to for example high action regularizations. Still, these context regions should be discovered, even if the auxiliary rewards $\tildegating$ are not sufficient. For this purpose, we calculate a value function, which reveals the mean reward in a context. If solely one component covers this context region, it will get a high value for updating its weight $\mgating$, although the task reward might be bad compared to other components in other regions. We use importance sampling, to calculate the Value function

\begin{align}\label{eq::appendix::val_func}
    V(\cvec c) = \sum_o\frac{\gatingpi}{\hat{\pi}(o|\cvec c)}\int_{\cvec \theta}\comppi \reward d\cvec \theta,
\end{align}

where $\hat{\pi}(o|\cvec c)$ are the gating probabilities before starting to update the weights $\mgating$ at the end of our algorithm (see Algorithm \ref{algo}). We completely resample the small replay buffer for all components before starting to update $\mgating$. We use the Nadaraya Watson predictor to get $V(\cvec c)$. Thus, the update rule for $\mgating$ is given by 
\begin{align}\label{eq::appendix::WeightUpdate}
    L_p &= \sum_o L_{c,k}(o) -V(\cvec c) + \beta \mathcal{H}\left(\mgating\right).
\end{align}

\section{Algorithmic Details}\label{sec::app:alg_details}
We describe algorithmic details in the algorithm box \ref{algo}. We start with one component, which is randomly added and incrementally add components after training the lastly added component for $K$ iterations. We update the component and context distributions individually. Note that there is no strict order on updating the components or context distributions first.

Every $H$ iterations we fine-tune all components, if more than one component is available. This allows the already trained components to adjust for the newly added one. After finishing training the lastly added component $k$, we check if it converged to a local optimum. There are different ways on checking that. For example by comparing the entropy, or achieved task reward to already trained components. This step is not absolutely necessary since for the upcoming fine tune steps, the component might improve. For the Beer Pong and Table Tennis task, we have disabled the deletion step in line 13.   

As last step we update the prior weights $\mgating$ and delete all components, which are below a threshold value (e.g. $10^{-5}$) for $\mgating$.

Note that in some experiments, it might happen that highly versatile skills result in worse rewards, e.g. through action regularization, although the task is solved successfully. In order to avoid deleting these diverse skills from the skill library at the end of the optimization, 
a higher $\beta $ value, $\beta_w$ for the entropy of $\mgating$ in Equation \ref{eq::appendix::WeightUpdate} can be used. This higher $\beta_w$ value will encourage to put weights on skills which are not achieving highest reward. Although those skills still will have lower weight compared to other components, they are kept in the skill set and might be useful. For example when the model should be adapted to an environmental change. We made use of a different $\beta_w$ value only once in the Beer Pong task (see Experimental Details \ref{sec::app:exp_details}). 

Note that we have used a variant of MORE \cite{abdolmaleki2016model} for optimizing obj. \ref{eq::LBCtxtDistr_with_aux}, as described in \cite{arenz2020trust} and a variant of Contextual MORE \cite{tangkaratt2017policy} for optimizing obj. \ref{eq::LBComps_without_aux} as described in \cite{Becker2020Expected}.

Also note, that for updating each component we use a small replay buffer. The newly taken samples replace the oldest samples in the buffer. For the updates we simply use all samples in the buffer for each component.

\begin{algorithm}[t]
	\caption{Versatile Skill Learning}
	\begin{algorithmic}[1]
		\renewcommand{\algorithmicrequire}{\textbf{Input:}}
		\renewcommand{\algorithmicensure}{\textbf{Output:}}
		\REQUIRE $\alpha, ~ \beta, ~\beta_w$, N, K, H
		\ENSURE  $\policy $
		\FOR{$k = 1$ to N}
		\STATE $\comppi, ~\statecomp \leftarrow $ randomly\_add\_component()
		\STATE $\mgating = 1/k, ~~ \forall{o} $
		\FOR{$ i = 1$ to $K$}
		\IF{i\% H is 0}
		    \STATE update\_all\_components($\alpha$) to obj. (\ref{eq::LBComps_without_aux}) each
		    \STATE update\_all\_context\_distributions($\alpha, ~ \beta$) to obj. (\ref{eq::LBCtxtDistr_with_aux}) each
	    \ELSE
		\STATE $\comppi \leftarrow$ update\_component\_k($\alpha$) to obj. (\ref{eq::LBComps_without_aux})
		\STATE $\statecomp \leftarrow$ update\_context\_distribution\_k($\alpha, ~ \beta$) to obj (\ref{eq::LBCtxtDistr_with_aux}) 
		\ENDIF
		\ENDFOR
		\IF{component\_k in local optimum}
		\STATE delete\_component\_k() 
		\ENDIF
		\ENDFOR
		\STATE $\mgating \leftarrow$ update\_prior\_weights($\beta_w$) to obj. (\ref{eq::appendix::WeightUpdate})
		\STATE delete\_redundant\_components()
	\end{algorithmic}
	\label{algo}
\end{algorithm}

\section{Experimental Details}\label{sec::app:exp_details}
We describe the different hyper parameters and environment specifications used in the experiments part here. Please note: We included a context punishment in all reward functions for all experiments for contexts which were sampled out of the context range defined by the environment. This encourages the context distributions $\statecomp$ to stay in the valid context region as defined from the environment. We report the reward functions and the environment specifications in the sub section for each experiment. 

As for the hyper parameters for all experiments for our Algorithm, we provide a summary in Table \ref{table::appendix_hyp_params_svsl}.
Following the notation from Algorithm \ref{algo}, $\beta_w$ is the entropy coefficient for $\mgating$ in Objective \ref{eq::appendix::WeightUpdate}, $N$ is the number of incrementally added components, $K$ is the number of iterations one component is trained and $H$ describes the number that every $H.$ iteration all components are updated.

For the hyper parameters for all experiments for HiREPS and LaDiPS, we provide a summary in Table \ref{table::appendix_hyp_params_hireps} and \ref{table::appendix_hyp_params_ladips} respectively.  

\begin{table}[h!]
    \begin{center}
        \begin{tabular}{|c|c|c|c|c|c|c|}
        \hline
         Hyper Parameter (Ours)    & $\alpha$ & $ \beta $ & $\beta_w$ & N & K & H \\\hline
         Planar Reacher (Ablation) & varies (see section \ref{sec::Ablation})        & 1.0         & 1.0         & 60 & 350 & 50\\\hline
         Beer Pong Task            & 0.001    & 0.5       & 2.5       & 70 & 750 & 50 \\\hline
         Table Tennis Task         & 0.00001        & 0.2         & 0.2         & 50 & 800 & 50 \\\hline
        \end{tabular}
            \caption{Hyperparameters of our algorithm for all environments.}
    \label{table::appendix_hyp_params_svsl}
    \end{center}
    \vspace{-8mm}
\end{table}

\begin{table}[h!]
    \begin{center}
        \begin{tabular}{|c|c|c|c|}
        \hline
         Hyper Parameter (HiREPS)  & $\epsilon$ & $ \kappa $ & N   \\\hline
         Planar Reacher (Ablation) & 0.5          & 0.99     & 60          \\\hline
         Beer Pong Task            & 0.3          & 0.9     & 70          \\\hline
         Table Tennis Task         & 0.3          & 0.9     & 50         \\\hline
        \end{tabular}
            \caption{Hyperparameters of HiREPS for all environments.}
    \label{table::appendix_hyp_params_hireps}
    \end{center}
    \vspace{-8mm}
\end{table}

\begin{table}[h!]
    \begin{center}
        \begin{tabular}{|c|c|c|c|c|}
        \hline
         Hyper Parameter (LaDiPS)  & $\epsilon$ & $ \epsilon_{gating} $ & $\alpha$ & N   \\\hline
         Beer Pong Task            & 0.1          & 0.05     &  $\log 30$  &  70          \\\hline
         Table Tennis Task         & 0.01          & 0.02     &  $\log 30$  &  50          \\\hline
        \end{tabular}
            \caption{Hyperparameters of LaDiPS.}
    \label{table::appendix_hyp_params_ladips}
    \end{center}
    \vspace{-8mm}
\end{table}

\subsection{Ablation Studies}
For the Planar Reaching task we consider a two dimensional context space. Note that we have not used any Movement Primitive representation for the policy. Instead, the sampled $\theta$ from our search distributions directly represent the angles of each joint of the reacher task. 

We consider a two dimensional context space $x, y$, where $x_{min} = 4.5, ~ x_{max} = 7$ and $y_{min}=-6, ~~ y_{max}=6$. 

The reward function is given by: 

\begin{align*}
    R(\cvec \theta, \cvec c) = -||\cvec \theta||^2_2-2\cdot||\cvec f(\cvec \theta) - \cvec c||^2_2 - l_{c}(\cvec c) -l_o(\cvec \theta), 
\end{align*}
where 
\begin{align*}
    l_{c}(\cvec c) =\left\{\begin{array}{ll} 10, & c \not\in C \\
         0, & else \end{array}\right., ~~~~
    l_{o}(\cvec c) =\left\{\begin{array}{ll} 3, & b_{1:10} \in O \\
         0, & else \end{array}\right. ,     
\end{align*}
where $l_c$ are the costs if a context $\cvec c$ is not within the valid context range $C$ and $l_o$ are the costs, if one of the ten links $b_i$ has a collision with the obstacles of the environment. $\cvec f(\cvec \theta)$ are the forward kinematics of the planar reacher. Note that the angles $\cvec \theta$ are always normalized into a range $[-\pi, \pi]$.

\subsection{Beer Pong}
For the Beer Pong task we consider a two dimensional context space $x, y$ positions of the cup on the table, where $x_{min}=-0.32, ~x_{max}=0.32$ and $y_{min}=-2.2, ~~y_{max}=-1.2$. We use ProMPs as policy parameterizations. We also learn the length of the trajectory, which might lead to invalid, i. e. negative trajectory length. We use a punishment term, if the trajectories are out of a reasonable range and do not execute this sample on the environment. We do the same for the context samples. Furthermore, we do not execute trajectories, which violate the joint constraints of the robot. Since these samples are not executed on the environment, they are not counted as "taken samples". For the case that those restrictions are not violated, the reward function is given as

\begin{align*}
    R(\cvec \theta, \cvec c) = \left\{\begin{array}{lll} -4 - min(||p_{c,top}-p_{b,1:T}||_2^2) - 0.5||p_{c,bottom}-p_{b,T}||_2^2 -10^{-4}\tau, & \textrm{if cond. 1} \\
                                                         -2 - min(||p_{c,top}-p_{b,1:T}||_2^2) - 0.5||p_{c,bottom}-p_{b,T}||_2^2 -10^{-4}\tau, & \textrm{if cond. 2} \\
                                                         ||p_{c,bottom}-p_{b,T}||_2^2 +1.5 \cdot 10^{-4} \tau, & \textrm{if cond. 3}
                                                         \end{array}\right.,
\end{align*}
where $p_{c, top}$ is the position of the top edge of the cup, $p_{c, bottom}$ is the ground position of the cup, $p_{b,t}$ is the position of the ball at time point $t$ and $\tau$ is the squared mean torque over all joints during one rollout. 
The different conditions are:
\begin{itemize}
    \item cond. 1: The ball is not in the cup and had no table contact
    \item cond. 2: The ball is not in the cup and had table contact
    \item cond. 3: The ball is in the cup.
\end{itemize}
For the case that context samples were sampled out of the range we provide a high negative reward
\begin{align*}
    R_c(\cvec \theta, \cvec c) = -30 - d_c^2,
\end{align*}
where $d_c^2$ is the distance of the current context $\cvec c$ to the valid context region.

For the case that the duration of the trajectory was sampled out of the range we provide a negative reward as
\begin{align*}
    R_{\theta}(\cvec c, \cvec \theta) = \left\{\begin{array}{ll} -30 -10\cdot(l-0.1)^2, & \textrm{if } l<0.1 \\
                                                                 -30 -10\cdot(l-1.3)^2, & \textrm{if } l>1.3 
                                                         \end{array}\right.,
\end{align*}
where $l$ is the duration of the Trajectory in seconds. If the joint constraint limits are violated by the ProMP's trajectory, we give a punishment of -30.

\subsubsection{On Versatility of our Solutions on the Beer Pong Task}\label{sec::apdx::VersatilityBP}

\begin{figure}[h!]
    \centering
   \resizebox{0.45\textwidth}{!}{\input{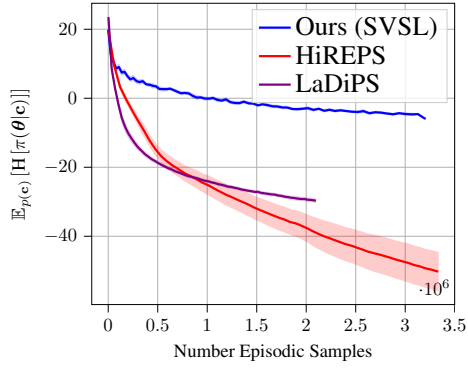}}
   \caption{Expected Mixture Entropies.}
   \label{fig::exps_bp_episodic::apdx:mixt_entr}
\end{figure}

In this section we want to compare the learned solutions of LaDiPS \cite{endLayered} and our method. As LaDiPS uses experience-sharing between components, it outperforms our method in terms of sample efficiency, but does not find as qualitative solutions as our method, which is reflected in the end-rewards (see Fig. \ref{fig::exps_bp_episodic::rewards}), where we can clearly achieve a higher value. In addition to the more qualitative solutions, we are able to learn a mixture model with higher expected entropy, which can be seen in Fig. \ref{fig::exps_bp_episodic::apdx:mixt_entr}). Our method is able to achieve more qualitative skills with a much higher entropy compared to LaDiPS and HiREPS, indicating that the proposed solutions by our method are more versatile than the others. Please note that we have defined a fine grid of contexts and sampled for each context $\cvec c$ 1000 parameter vectors $\cvec \theta$ from the mixture models to calculate the expected entropy. We have repeated this procedure for all 20 different trials/seeds and report the mean expected entropy together with two times the standard error. 
In addition to the expected mixture entropy we have conducted a qualitative comparison, where we have picked the first model for both, our method and LaDiPS and have let it run for twelve different contexts. For each context we have sampled 100 times from the mixture model and executed the mean of the sampled components on the environment. Note that we only have taken those ball trajectories into account that lead to a successful solution. For LaDiPS all observed solutions for all contexts showed only one mode, in which the ball bounced once on the table and then landed in the cup. For our method the same solution could be seen for six contexts. For three contexts, our method threw the ball directly into the cup, whereas for another three contexts three different ball trajectories could be observed: i) direct throw, ii) bouncing ones on the table ii) bouncing once on the table and once on the wall.

\subsection{Table Tennis}
For the Table Tennis task we consider a four dimensional context space $[\cvec{b^{des}, ~b^{inc}}] \in\mathbb{R}^4$. We have the desired landing position of outgoing ball $ \cvec{b^{des}} = [x^{des}, ~y^{des}]\in\mathbb{R}^2$  and the initial positions of the incoming ball $ \cvec{b^{inc}} = [x^{inc}, ~y^{inc}]\in\mathbb{R}^2$ where 
$x^{des}\in[-1.2, ~-0.2], ~y^{des}\in[-0.6, ~0.6], ~x^{inc}\in[-1.2, ~-0.2],
~y^{inc}\in[-0.65, ~0.65].$
We fix the initial ball velocity and  $z^{inc}$ position so we can transform the initial ball position to the incoming ball landing position on the table. We use ProMPs as policy parameterizations and learn the length of the trajectory, which might lead to invalid, i. e. negative trajectory length. We use a punishment term, if the length of trajectories are out of a reasonable range and do not execute this sample on the environment. We do the same for the context samples. Furthermore, we do not execute trajectories, which violate the joint constraints of the robot and return a punishment. Since these samples are not executed on the environment, they are not counted as "taken samples". For the case that those restrictions are not violated, the reward function is given as

\begin{align*}
    R(\cvec \theta, \cvec c) = \left\{\begin{array}{lll} 0.2 \cdot(1 - tanh(  \min\limits_{t}(||\cvec {b_{t}}-\cvec{r_{t}}||_2^2)), & \textrm{if cond. 1} \\
                                                         2 \cdot(1-tanh(  \min\limits_{t}(||\cvec {b_{t}}-\cvec{r_{t}}||_2^2)) \\  + ~ (1-tanh(\min\limits_{t}(||\cvec{b^{des}}-\cvec {b_{t,xy}}||_2^2)) , & \textrm{if cond. 2} \\
                                                          2 \cdot  (1-tanh(\min\limits_{t}(||\cvec {b_{t}}-\cvec{r_{t}}||_2^2))  \\ +~ 4\cdot(1-tanh((||\cvec{b^{des}}-\cvec{b^{land}}||_2^2)) + 1_{\{ b^{land}_{x} < 0\}}, & \textrm{if cond. 3}
                                                         \end{array}\right.,
\end{align*}
where $\cvec {b_{t}}\in\mathbb{R}^3$ is the position of the ball at time point $t$, $ \cvec {b_{t,xy}}\in\mathbb{R}^2$ is the $x,y$ position of the ball at time point $t$, $\cvec{r_{t}}\in\mathbb{R}^3$ is the position of the racket at time point $t$, $\cvec{b^{land}} \in\mathbb{R}^2$ is the real $x,y$ landing position point. 
The different conditions are:
\begin{itemize}
    \item cond. 1: The ball had no racket contact.
    \item cond. 2: The ball had racket contact but then no table contact.
    \item cond. 3: The ball had racket contact and then table contact. 
\end{itemize}
For the case that context samples were sampled out of the range we provide a high negative reward
\begin{align*}
    R_c(\cvec \theta, \cvec c) = -20 \cdot d_c^2,
\end{align*}
where $d_c^2$ is the distance of the current context $\cvec c$ to the valid context region.

For the case that the duration of the trajectory was sampled out of the range we provide a negative reward as
\begin{align*}
    R_{\theta}(\cvec c, \cvec \theta) = \left\{\begin{array}{ll} -3||l-0.1||, & \textrm{if } l<0.1 \\
                                                                 -3||l-5||, & \textrm{if } l>5 
                                                         \end{array}\right.,
\end{align*}
where $l$ is the duration of the Trajectory in seconds. 

For the case that the trajectory was sampled out of the joint constraints we provide a negative reward as
\begin{align*}
    R_c(\cvec \theta, \cvec c) = -1 \cdot \frac{1}{T} \sum\limits_{i}^{T}d_{qpos,t},
\end{align*}
where $d_{qpos,t}$ is the distance of the planned joint positions to the valid joint position region.

\subsubsection{On Versatility of our Solutions on the Table Tennis Task}
\begin{figure}[h!]
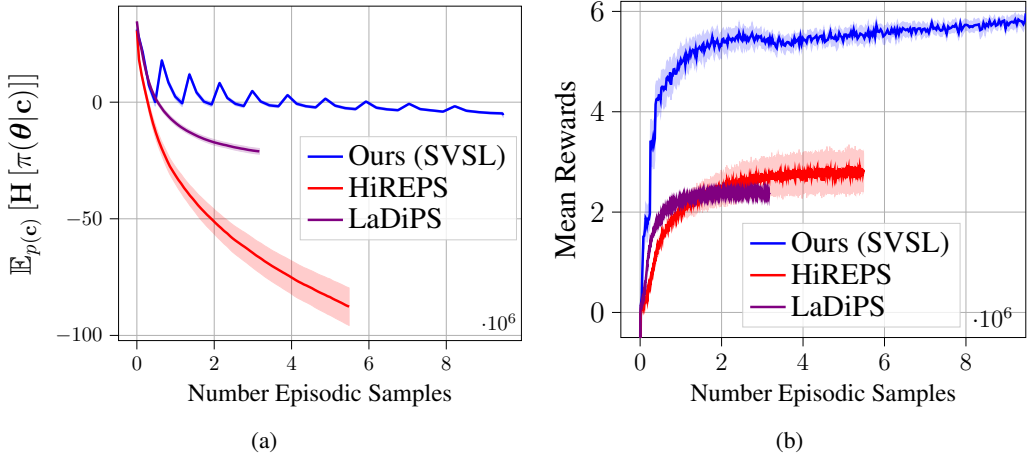

    \begin{minipage}[b]{0.5\textwidth}
       \centering
       \resizebox{1\textwidth}{!}{\input{plots/experiments/episodic_tt_small_range/exp_mixture_entropies}}
       \subcaption[]{}
       \label{fig::exps_tt_episodic::apdx:mixt_entr}
   \end{minipage}\hfill
   \begin{minipage}[b]{0.5\textwidth}
        \centering
       \resizebox{0.92\textwidth}{!}{\input{plots/experiments/episodic_tt_small_range/rewards_with_ladips_long}}
       \subcaption[]{}
       \label{fig::exps_tt_episodic::apdx:long_reward}
    \end{minipage}\hfill%
    \caption{\textbf{Expected Mixture Entropies} (left) and \textbf{performance on the table tennis task} (right). Note that the performance graph is the same as in Fig. \ref{fig::exps_tt_episodic::rewards}, but with the full graph of our method to train 70 components. The graph in Fig. \ref{fig::exps_tt_episodic::rewards} was clipped in the x-axis to maintain overview.}
\end{figure}

We compare the versatility of learned solutions of LaDiPS \cite{endLayered}, HiREPS \cite{daniel2012hierarchical} and ours. We plot the expected mixture entropies of each algorithm in Fig. \ref{fig::exps_tt_episodic::apdx:mixt_entr}. For each of the 1600 uniformly context samples we have sampled 1000 parameter vectors $\boldsymbol{\theta}$ from the mixture model to calculate the mixture entropy. We repeat this for all models resulting from 20 seeds/trials and plot the average of the entropies in Fig. \ref{fig::exps_tt_episodic::apdx:mixt_entr}. Please note that the fluctuations of the entropy curve in Fig. \ref{fig::exps_tt_episodic::apdx:mixt_entr} of our algorithm are due to adding components during training.

We are able to achieve the highest entropy while clearly outperforming both methods in terms of the achieved task reward (see Fig. \ref{fig::exps_tt_episodic::apdx:long_reward}), which indicates that we are able to learn more qualitative skills.

\subsection{A Comparison Between Episodic- and Step-Based Policy Search}\label{AppdxCompStepBasedEpisodic}
The step-based and the episodic policy search setting have their own benefits and limitations. 
For example, while in the step-based case we usually explore the action space using uncorrelated noise, the episodic setting uses correlated distributions directly in the parameter space of the controller \cite{deisenroth2013survey}. Yet, step-based approaches can use the data more efficiently as they decompose the episodes into single time steps. Hence, it is interesting to compare these approaches.
\begin{minipage}{0.55\textwidth}
    We compare the performance of our algorithm against PPO, a well-known Deep Reinforcement Learning method \cite{schulman2017proximal}. 
    We want to analyze when episodic exploration is beneficial over step-based exploration and vice-versa for different sparsity levels of the reward.  
    For this purpose, we use the Reacher task from OpenAI gym \cite{openaigym}. In this task, the agent has to reach a goal position with its tip. In order to consider a more difficult task, we extend the original reacher set-up and show the considered environment in Figure \ref{fig::appdx::reacher_setup}). First, we increase the number of links from two to five and we fix the initial state position to be initialized on the horizontal to the left (see Fig. \ref{fig::appdx::reacher_setup}), instead of initializing on the horizontal to the right side as it is in the original version. 
\end{minipage}\hfill
\begin{minipage}{0.4\textwidth}
\begin{figure}[H]
    \centering
    \includegraphics[width=0.98\textwidth]{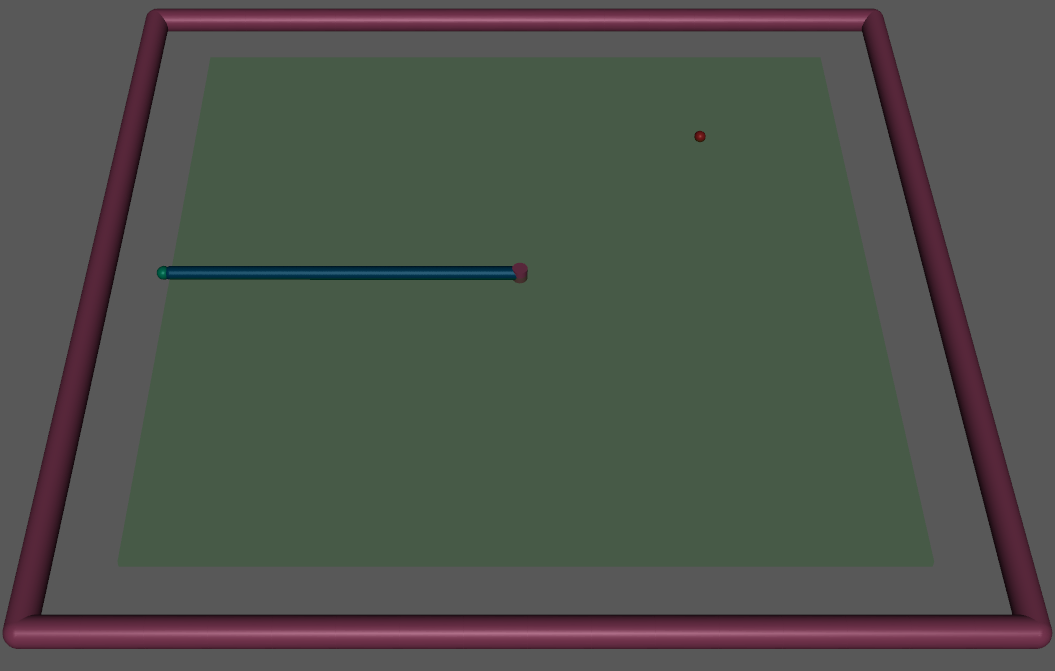}
    \caption{5Link Reacher Set-up. The reacher's tip needs to reach the red goal position. Adapted from \cite{openaigym}.}
    \label{fig::appdx::reacher_setup}
\end{figure}
\vspace{2mm}
\end{minipage}
Furthermore, we fix the initial joint velocity to be zero. For the original 2-Link case, the goal reaching space is the circle around the agent's base, whereas in our case we consider the right half-circle around the agent's base, where the radius is the agent's total length. By initializing the agent on the left side and only considering goal positions on the right side, we make the task harder. We compare our method to the popular PPO algorithm \cite{schulman2017proximal} and use the implementation from \cite{stable-baselines3}.

In order to examine the different policy search strategies we consider three different markovian reward functions in the same set-up. The first is given by  
\begin{align}\label{originalReward}
    R_t(\cvec s, \cvec a) = -||\cvec t(\cvec s) - \cvec g||_2 - \sum_ia_i^2,
\end{align}
where $\cvec s$ is the current state, $\cvec a$ is the current action, $t(\cvec s)$ is the tip's current position and $\cvec g$ is the goal position. Except for $\cvec g$, which is re-sampled at each episode, the values are given at the current time-step $t$. One episode lasts for $T=200$ time-steps.

This reward function is also given by \cite{openaigym} and encodes task information, i.e. tip distance to the goal, in each time-step. In the following we only change the reward function. 

The second reward function is given as
\begin{align}\label{mediumSparseReward}
    R_{t}(\cvec s, \cvec a) = \left\{\begin{array}{ll}  - \sum_i a_i^2, & \textrm{if } t<190 \\
                                                        -||\cvec t(\cvec s) - \cvec g||_2 - \sum_i a_i^2, & \textrm{if } t\geq 190 
                                                         \end{array}\right..
\end{align}

Characteristic for this reward function is that it will only return task information, i.e. tip-distance to goal, in the last 10 steps of the episode. 

The third reward function is given as 
\begin{align}\label{sparsestSparseReward}
    R_{t}(\cvec s, \cvec a) = \left\{\begin{array}{ll}  - \sum_i a_i^2 - \sum_i v_i^2, & \textrm{if } t<198 \\
                                                        -500\cdot||\cvec t(\cvec s) - \cvec g||_2 - \sum_i a_i^2 - 1000\cdot\sum_i v_i^2, & \textrm{if } t\geq 198
                                                         \end{array}\right.,
\end{align}
where $\cvec v$ is the joint velocity at time-step $t$. This reward function is very similar to the reward function from Eq. (\ref{mediumSparseReward}). However, it only returns task information in the last two steps of the episode, which makes the task very difficult. 

The reward functions in Equations (\ref{mediumSparseReward}, \ref{sparsestSparseReward}) can be motivated from the optimal control point of view. There, usually controllers with minimum energy consumption are seeked giving rise to punishments to taken control actions in the first steps as it is done by the proposed reward functions in Equations (\ref{mediumSparseReward}, \ref{sparsestSparseReward}).  

For PPO we use the same observation space as described in \cite{openaigym}, but for the Objectives (\ref{mediumSparseReward}, \ref{sparsestSparseReward}) we augment this observation space with the current time-step $t$.

The context $\cvec c$ for our method is given through the two-dimensional goal position vector $\cvec g$. 

\begin{figure}[t!]
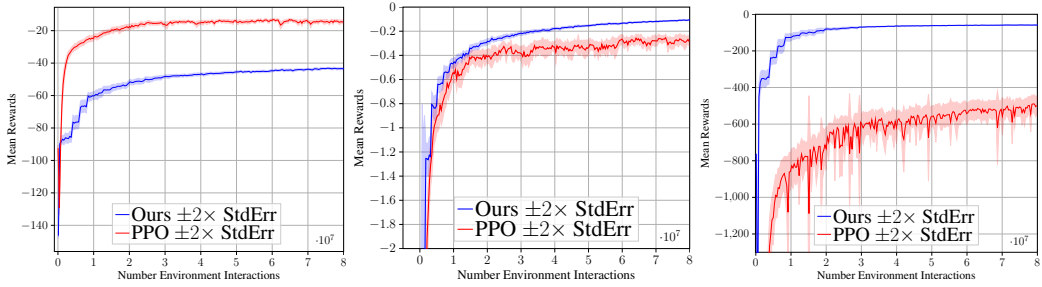

    \begin{minipage}[b]{0.33\textwidth}
        \centering
       \resizebox{\textwidth}{!}{\input{plots/experiments/EP_VS_STEP/norm_env}}
       \subcaption[]{Performances for Reward (\ref{originalReward})}
       \label{fig::appdx_comp::rewards_normal}
   \end{minipage}%
    \begin{minipage}[b]{0.33\textwidth}
        \centering
       \resizebox{\textwidth}{!}{\input{plots/experiments/EP_VS_STEP/medium_env}}
       \subcaption[]{Performance for Reward (\ref{mediumSparseReward})}
       \label{fig::appdx_comp::rewards_medium}
   \end{minipage}%
           \begin{minipage}[b]{0.33\textwidth}
       \centering
       \resizebox{\textwidth}{!}{\input{plots/experiments/EP_VS_STEP/sparsest_env}}
       \subcaption[]{Performance for Reward (\ref{sparsestSparseReward})}
       \label{fig::appdx_comp::rewards_sparsest}
   \end{minipage}
   \caption{\textbf{A Comparison of Episodic Policy Search to Step-based Policy Search.} We compare our episodic policy search method to the popular step-based approach PPO. We analyze the performance on three different reward types. For (a) we consider the reward in Equation (\ref{originalReward}). Here, task-information are returned in each time-step. For(b) we consider the reward in Equation (\ref{mediumSparseReward}). Here, task-information are returned in the last 10 steps of the episode. For (c) we consider the reward in Equation(\ref{sparsestSparseReward}), where only at the last 2 time-steps reward information are returned.}
   \label{fig::appdx:comparison}
	\vspace{-1.7mm}
\end{figure}

We report the average performance and two times the standard error over 20 seeds/trials for each experiment in Fig. \ref{fig::appdx:comparison}). The performances for the reward function in Equation (\ref{originalReward}) are shown in Fig. \ref{fig::appdx_comp::rewards_normal}). Clearly, PPO outperforms our algorithm, indicating that task information in each time-step helps exploring in the raw action space to find solutions. The performances for the reward function in Equation  (\ref{mediumSparseReward}) is shown in Fig. \ref{fig::appdx_comp::rewards_medium}). While PPO also performs well, we can observe that our algorithm can make use of the exploration in the parameter space and outperform PPO. This effect can be seen much more clearly for the reward funcion in Equation (\ref{sparsestSparseReward}). Fig. \ref{fig::appdx_comp::rewards_sparsest}) shows that PPO has clear problems to solve this task. For the settings given by the reward functions in Equations (\ref{mediumSparseReward}, \ref{sparsestSparseReward}) exploration in the parameter space as done by episodic policy search methods lead to better performance, whereas exploration in the raw action space leads to converging to a local optimum.

\subsubsection{Hyperparameters}

The hyperparameters for our algorithm are given in the following Table. Please note that we used punishments,

\begin{table}[h!]
    \begin{center}
        \begin{tabular}{|c|c|c|c|c|c|c|}
        \hline
        Hyper Parameter (Ours)                      & $\alpha$ & $ \beta $ & $\beta_w$ & N & K & H \\\hline
        Reacher Reward (\ref{originalReward})       &  $1e^{-6}$&  15      &  15       & 40 & 250 & 50\\\hline
        Reacher Reward (\ref{mediumSparseReward})   & 0.01    & 1          & 2.5       & 40 & 250 & 50 \\\hline
        Reacher Reward (\ref{sparsestSparseReward}) &  1      & 25         & 25        & 40 & 800 & 50 \\\hline
        \end{tabular}
    \caption{Hyperparameters of our algorithm for the reacher environment.}
    \label{table::appendix_hyp_params_reacher_svsl}
    \end{center}
    \vspace{-8mm}
\end{table}
if contexts were sampled out of the context range (which is the right half circle around the base with radius 0.5), we provided a quadratic punishment to the distance of the sample to the valid context bound. We multiplied this distance with a factor of 1000 for the case of reward function (\ref{originalReward}, \ref{mediumSparseReward}) and with a factor of 50000 for the case of reward function (\ref{sparsestSparseReward}).  

We normalized the observations and rewards for PPO and used the same hyperparameters for all three experiments, which we summarized in the following table

\begin{table}[h!]
    \begin{center}
        \begin{tabular}{|c|c|c|c|c|c|c|c|}
        \hline
        Hyp. Params (PPO)                      & $\alpha$ & N & GAE $\lambda$ & $\gamma$ & K & $\epsilon$ & network structure\\\hline
        Reacher Reward (\ref{originalReward})       & $1e^{-4}$ & 16384 &  0.95 & 0.99 & 64 & 0.2& [64, 64]\\\hline
        Reacher Reward (\ref{mediumSparseReward})   & $1e^{-4}$ & 16384 &  0.95 & 0.99 & 64 & 0.2& [64, 64]\\\hline
        Reacher Reward (\ref{sparsestSparseReward}) & $1e^{-4}$ & 16384 &  0.95 & 0.99 & 64 & 0.2& [64, 64]\\\hline
        \end{tabular}
    \caption{Hyperparameters of PPO for the reacher environment. Note that $\alpha$ is the learning rate, N is the number of rollouts, $\gamma$ is the discount factor, K is the batch size, $\epsilon$ is the clip value for the importance ratio. The layer structure is for both: the policy network as well as the value function network. }
    \label{table::appendix_hyp_params_reacher_ppo}
    \end{center}
    \vspace{-8mm}
\end{table}
\end{document}